\documentclass[journal]{IEEEtran}

\ifCLASSINFOpdf
\else
\fi
\usepackage{cite}
\usepackage{amsmath,amssymb,amsfonts}
\usepackage{algorithmic}
\usepackage{graphicx}
\usepackage{booktabs}
 \usepackage{multirow}
\usepackage{textcomp}
\usepackage{xcolor}
\usepackage[table]{xcolor}
\usepackage{bm}

\begin{document}
\bstctlcite{IEEEexample:BSTcontrol} 

\title{Adaptive Factor Graph-Based Tightly Coupled GNSS/IMU Fusion for Robust Positioning}

\author{Elham~Ahamdi, 
        Alireza~Olama, 
        Petri~Välisuo,~\IEEEmembership{Member,~IEEE,} 
        and~Heidi~Kuusniemi,~\IEEEmembership{Member,~IEEE}
\thanks{Elham Ahmadi, and Petri Välisuo are with the School of Technology and Innovations, University of Vaasa, Wolffintie 32, 65200, Vaasa, Finland e-mail: (elham.ahmadi@uwasa.fi, petri.valisuo@uwasa.fi).}
\thanks{Alireza Olama is with Information Technology, Åbo Akademi University, Vaasa, Rantakatu 2, 65100, Vaasa, Finland e-mail: (alireza.olama@abo.fi).}
\thanks{Heidi Kuusniemi is with the Faculty of Information Technology and Communication Sciences, Tampere University, Korkeakoulunkatu 1, 33720, Tampere, Finland, and also with Finnish Geospatial Research Institute, Espoo, Finland e-mail: (heidi.kuusniemi@tuni.fi).}
}


\maketitle

\begin{abstract}
Reliable positioning in GNSS-challenged environments remains a critical challenge for navigation systems. Tightly coupled GNSS/IMU fusion improves robustness but remains vulnerable to non-Gaussian noise and outliers. We present a robust and adaptive factor graph-based fusion framework that directly integrates GNSS pseudorange measurements with IMU preintegration factors and incorporates the Barron loss, a general robust loss function that unifies several m-estimators through a single tunable parameter. By adaptively down weighting unreliable GNSS measurements, our approach improves resilience positioning. The method is implemented in an extended GTSAM framework and evaluated on the UrbanNav dataset. The proposed solution reduces positioning errors by up to 41\% relative to standard FGO, and achieves even larger improvements over extended Kalman filter (EKF) baselines in urban canyon environments. These results highlight the benefits of Barron loss in enhancing the resilience of GNSS/IMU-based navigation in urban and signal-compromised environments.
\end{abstract}


\IEEEpeerreviewmaketitle


\section{Introduction}
\IEEEPARstart{R}{eliable} and accurate navigation is paramount for a vast array of applications, from autonomous vehicles and robotics to surveying and personal navigation devices. Global navigation satellite systems (GNSS) and inertial measurement units (IMUs) are two cornerstone sensor modalities that, when fused, provide a robust and continuous navigation solution \cite{groves2015principles, soloviev2010tight}.
GNSS encompasses various satellite constellations, including the United States' GPS, Russia's GLONASS, Europe's Galileo, and China's BeiDou. These systems operate based on a network of satellites orbiting the Earth, a ground control segment for monitoring and corrections, and user receivers \cite{Teunissen_2017}. GNSS signals are susceptible to various error sources, including atmospheric delays (ionospheric and tropospheric), multipath propagation (signal reflection off surfaces like buildings), non-line-of-sight (NLOS) reception, satellite orbital and clock errors, receiver noise, and intentional or unintentional interferences (jamming and spoofing) \cite{zidan2020gnss}. These factors introduce significant noise and outliers into the GNSS measurements.
To address these challenges, research has increasingly focused on developing resilient positioning solutions.

An IMU, on the other hand, is a self-contained sensor package that measures a body's specific force and angular rate using a combination of accelerometers and gyroscopes, respectively, independent of external signals \cite{woodman2007introduction}. 
A critical characteristic of IMUs is that their measurements are subject to errors such as biases (constant or slowly varying offsets), scale factor errors, non-orthogonality of axes, and random noise \cite{groves2015principles}. These errors, particularly biases, cause the navigation solution derived from standalone IMU integration to drift over time, with position errors typically growing cubically and velocity errors quadratically in the absence of external corrections.

GNSS and IMU technologies possess highly complementary characteristics, making their fusion a powerful strategy for navigation \cite{miller2012sensitivity}.
GNSS provides absolute position and velocity information that is generally stable over the long term, but its signals are prone to blockage or severe degradation. Conversely, an IMU provides continuous, high-frequency measurements of relative motion and is immune to external interference or signal blockage. However, the integration of IMU measurements leads to the accumulation of errors.
GNSS measurements can be used to periodically correct drift errors inherent in the IMU solution, including estimating and compensating for IMU sensor biases. In turn, the IMU can bridge periods of GNSS outage or degradation, providing a continuous navigation solution. The fusion results in a navigation system with improved accuracy, availability, and robustness compared to either system operating standalone.

The integration of GNSS and IMU data can be achieved through different coupling strategies, most notably including loosely coupled (LC)
and tightly coupled (TC)
approaches, among others \cite{wen2021factor}. In an LC architecture, the GNSS receiver processes satellite signals independently to compute its own position and velocity estimates, which are then fused with the IMU solution. LC systems are simpler to implement but limited by their dependence on a complete GNSS solution.
In a TC architecture, raw GNSS measurements such as pseudoranges, Doppler shifts, and carrier phase measurements are directly integrated with IMU data. This enables advanced error modeling and higher potential accuracy than LC systems, but they are generally more complex to implement.

\begin{figure}[tb!]
    \centering
    \includegraphics[width=0.8\linewidth]{ 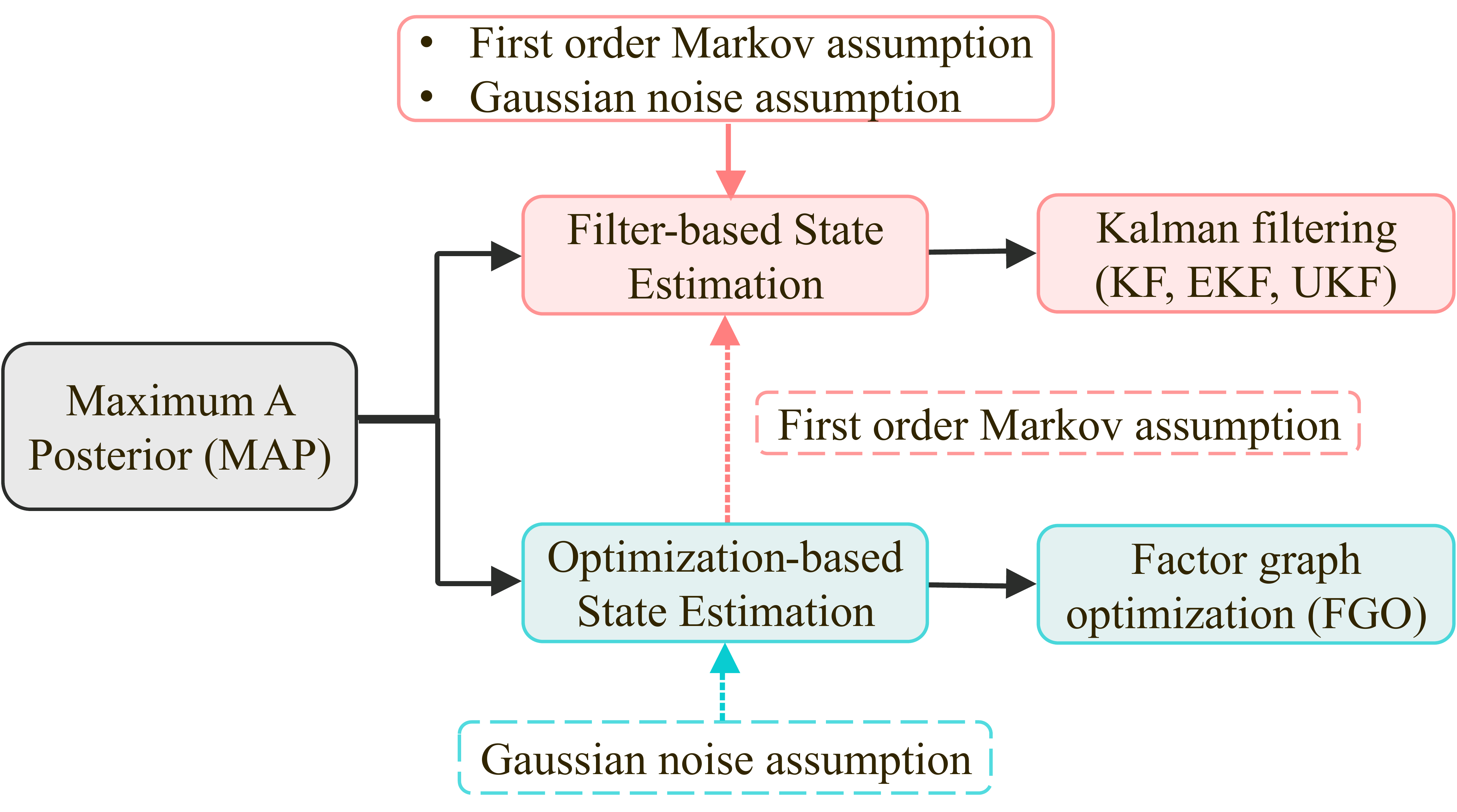}
    \caption{State estimation in sensor fusion: filter-based vs. optimization-based.
    }
    \label{fig_MAP_estimation}
\end{figure}

Sensor fusion and integration in modern perception and navigation systems are fundamentally concerned with estimating the system state from noisy, multi-modal sensor data. This estimation problem is commonly formulated within a Bayesian framework, where the goal is to compute the maximum a posteriori (MAP) estimate of the state \cite{murphy2012machine}. As illustrated in Fig.~\ref{fig_MAP_estimation}, the MAP estimation problem can be approached through two main categories: filter-based methods, which recursively estimate the state over time, such as Kalman filter (KF) and its variants, assuming a first-order Markov process and Gaussian noise \cite{crassidis2006sigma}, and optimization-based methods, which solve a global inference problem using techniques like factor graph optimization (FGO) \cite{dellaert2017factor}. Both approaches are widely adopted in robotics, autonomous systems, and geospatial applications.
However, most real-world navigation systems, including GNSS/IMU integration, involve non-linear dynamics and measurement models.
The extended Kalman filter (EKF) addresses this by linearizing these models around the current state estimate at each step. This linearization, however, introduces errors, especially if the non-linearities are significant or the state estimate is poor. Additionally, the performance of KFs can degrade significantly if the noise characteristics deviate from the assumed Gaussian distribution, which is often the case with GNSS measurements in urban environments due to multipath and NLOS. These limitations have spurred research into alternative estimation frameworks.

Factor graph optimization (FGO) has emerged as a compelling alternative to traditional filtering methods for state estimation \cite{dellaert2017factor}.
FGO is a batch estimation technique that formulates the state estimation problem as a nonlinear least squares (NLS) optimization over a graph structure. This graph consists of nodes representing the unknown state variables at different time steps and factors representing probabilistic constraints imposed by sensor measurements or motion models. 
FGO was initially popularized in the field of simultaneous localization and mapping (SLAM) in robotics, where it demonstrated significant advantages in terms of accuracy and robustness for estimating robot trajectories and environmental maps. Its application has since expanded to various navigation problems, including GNSS/IMU fusion.
Unlike EKF, which fixes linearizations
and marginalizes past states, FGO improves accuracy by optimizing over full trajectories, re-linearizing iteratively to reduce non-linear errors, and handling outliers with robust cost functions. It better exploits temporal correlations and offers modular, flexible integration of multiple sensors.

Despite the significant advances enabled by FGO in GNSS/IMU fusion, robust performance in the presence of measurement outliers and non-Gaussian noise remains challenging. Traditional robust optimization techniques often rely on the m-estimator loss functions such as Huber, Cauchy, or Tukey \cite{ben2024robust}, which attenuate the influence of outliers by down weighting residuals beyond a certain threshold \cite{ ahmadi2025robust}. However, these losses require manual tuning of scale parameters and make strong assumptions about the distribution of residuals. They also lack adaptivity to dynamic error characteristics that vary across time and environments, such as abrupt signal loss in urban canyons.
Addressing these limitations is essential for resilient positioning in real-world navigation with sensor degradation, disturbances, and dynamic noise.

In this paper, we propose a robust tightly coupled GNSS/IMU fusion framework based on adaptive FGO that integrates the general and adaptive robust loss function called Barron loss, introduced in \cite{barron2019general}, for enhancing resilient positioning.
The Barron loss function offers a principled, general formulation that encompasses a wide range of m-estimators within a single, continuous, and differentiable family of robust loss functions. Crucially, it introduces a shape parameter that can be adaptively tuned to optimize the trade-off between robustness and efficiency, thus mitigating the need for heuristic parameter selection \cite{de2021review}. By leveraging this adaptivity within a tightly coupled GNSS/IMU FGO framework, it becomes possible to dynamically suppress measurement anomalies while preserving the fidelity of reliable data. This capability is particularly beneficial in environments where error characteristics evolve rapidly, such as during transitions between open-sky and GNSS-denied areas, enabling more reliable state estimation and enhanced navigation resilience.
To evaluate the performance of the proposed method, we conduct experiments using real-world UrbanNav dataset \cite{hsu2021urbannav}, which serves as established benchmark in autonomous navigation.

The key contributions of this work are as follows.
First, we formulate a TC GNSS/IMU fusion framework within robust FGO architecture, where raw GNSS pseudorange measurements are directly fused with IMU preintegration factors. While TC FGO-based fusion has been explored in prior works, our formulation emphasizes robustness to real-world signal degradation for addressing GNSS vulnerabilities through an architecture explicitly designed for seamless integration with adaptive robust loss functions.
Second, we incorporate the use of a general and adaptive robust loss function, Barron loss, into the optimization process,
which generalizes several widely used m-estimators through a single tunable parameter. This formulation enables a smooth and continuous transition between different robustness levels, allowing the optimization process to adaptively adjust to varying noise levels and outlier characteristics in the measurements.
Third, we extend the open-source GTSAM library \cite{gtsam2022} by implementing the Barron loss as a built-in robust loss model, facilitating reproducibility and integration with existing factor graph pipelines.
Fourth, we demonstrate through simulation and real-world experiments that our proposed approach outperforms conventional FGO methods, including those employing fixed m-estimator robust losses, and also EKF fusion. 
Experimental results demonstrate improvements in terms of both positioning accuracy and robustness, particularly under GNSS-degraded conditions. To the best of our knowledge, this is the first application of the Barron loss in GNSS/IMU fusion, offering a novel and adaptive approach to robust optimization for resilient navigation.

This paper is structured as follows.
Section \ref{related work} reviews the development of factor graphs, their application to GNSS/IMU fusion, and recent advances in robust integration methods.
Section \ref{BACKGROUND} presents the necessary mathematical background, including frame definitions, MAP estimation, and robust statistics.
Section \ref{formulation} formulates the GNSS/IMU integration using FGO and introduce an adaptive robust optimization approach.
Section \ref{Results} provides experimental results that validate the proposed method and finally
Section \ref{conclusion} summarizes the findings and outlines future research directions.

\section{LITERATURE REVIEW} \label{related work}
The transition from sequential filtering methods like the Kalman filter to batch optimization techniques like FGO marks a significant evolution in addressing complex state estimation challenges in navigation.

\subsection{Early Developments in Factor Graphs}
The concept of representing estimation problems as graphs of constraints is not new. The seminal work by Lu and Milios \cite{lu1997globally} introduced the idea of globally consistent scan matching for map building, which involved representing robot poses and their relative constraints in a graph structure. Although not explicitly termed "factor graphs" at the time, their approach captured the essence of formulating SLAM as an optimization problem over a graph of pose relationships.

The formal introduction and popularization of factor graphs for SLAM, and more broadly for robotic perception, gained significant momentum with the work of Dellaert and Kaess \cite{dellaert2006square}. They explicitly framed the full SLAM problem as an NLS problem on a factor graph. 
Thrun and Montemerlo \cite{thrun2006graph} presented the GraphSLAM algorithm, which also unified optimization techniques for offline SLAM. These works established FGO as the prevailing framework for batch SLAM, enabling its extension to estimation problems such as GNSS/IMU fusion.

\subsection{Application of Factor Graphs to GNSS/IMU Fusion}
The proven success of FGO in SLAM has naturally extended its application to state estimation through multi-sensor fusion. This area has been extensively studied in the literature, with several review articles and tutorials consolidating the existing knowledge. \cite{dellaert2017factor} published a foundational text on factor graphs for robot perception, though more general than just GNSS/IMU, that explains the principles of FGO in robotics.
\cite{taylor2024factor} provided a comprehensive tutorial on factor graphs for navigation applications, highlighting their advantages over traditional filtering methods such as the Kalman filter.
The paper systematically introduces the mathematical formulation of factor graphs and demonstrates their application in modeling sensor measurements, including GNSS/IMU data. Emphasis is placed on the flexibility of factor graph optimization frameworks to handle nonlinearities, incorporate robust loss functions, and support modular sensor fusion.

In the context of LC and TC GNSS/IMU fusion, FGO offers significant advantages \cite{wen2021factor}, especially due to its capability for batch optimization and effective management of system nonlinearities.
The success of FGO in real-world scenarios, notably in Google's Smartphone Decimeter Challenge, highlights its capability to handle non-linearities and highly corrupted measurements effectively \cite{siemuri2024optimal}.
Specialized FGO libraries with GNSS-specific factors, such as gtsam-gnss \cite{suzuki2025open}, further facilitated research and application in this domain. These libraries provide implementations for pseudorange, Doppler, and various carrier phase factors, along with handling of ambiguities.
Early work by Indelman et al. \cite{indelman2013information} demonstrated the efficacy of factor graphs for fusing IMU, GNSS, and
vision measurements, leveraging incremental smoothing to achieve robust multi-sensor integration, laying the foundation for subsequent research in TC GNSS/IMU fusion.

A comprehensive comparison by Wen et al. \cite{wen2021factor} evaluated FGO against EKF for GNSS/INS integration using real-world data collected in an urban canyon in Hong Kong.
They demonstrated improved accuracy over EKF.
\cite{chai2025optimization} investigated
the LC GNSS/INS combination models based on EKF, robust KF, and FGO algorithms to enhance robust positioning in urban environments utilizing both high- and low-cost devices. Experimental results demonstrate that FGO outperforms EKF and RKF in position, velocity, and attitude estimation, particularly in scenarios of simulated gross errors and loss of GNSS signal.
Shen \textit{et.al} \cite{shen2024novel} proposed a TC GNSS/INS framework with carrier-phase ambiguity resolution, using a sliding window optimizer to fuse multi-GNSS pseudorange and carrier-phase observations with inertial measurements.
Several works have explored the integration of additional sensors  with GNSS/IMU in FGO framework to enhance localization accuracy. One method combines GNSS, LiDAR, and IMU using a two-stage factor graph optimization \cite{liu2023glio}, another approach fuses GNSS, visual, and IMU data and employed a coarse-to-fine initialization. \cite{zhang2024gnss} integrated GNSS, LiDAR, IMU, and optical speed sensors, used Gaussian process regression to enable time-centric graph construction without strict synchronization.

\subsection{Robust Methods in Factor Graph-Based GNSS/IMU Fusion}
To enhance resilience to GNSS signal degradation, several robust techniques have been proposed. 
Li \textit{et.al} \cite{li2018robust} presented a robust graph-based optimization method for TC GNSS/INS integrated navigation. By introducing a reliability factor into the factor graph, the method can down-weight faulty GNSS measurements. The approach is tested on simulated data with artificial noise, and while results show improved accuracy and fault detection, real-world validation is left for future work.
Atia \textit{et.al} \cite{atia2019map} introduced a map-aided adaptive fusion scheme that uses digital maps to detect and mitigate GNSS errors in urban environments, improving positioning under multipath and NLOS conditions.
Hu \textit{et.al} developed in \cite{hu2024robust}, a robust FGO-based method for maritime applications, incorporating factors based on pseudorange residual judgment to handle occluded GNSS signals, ensuring continuous and accurate navigation.
Girrbach \textit{et.al} presented a moving horizon estimation framework for GNSS/IMU fusion in \cite{girrbach2017optimization}, which dynamically adjusts the optimization horizon based on new measurement data, achieving improved state estimation accuracy.

In recent developments, robust estimation techniques have been integrated into FGO to enhance the resilience of GNSS/IMU fusion systems against outliers and non-Gaussian noise, particularly in urban environments \cite{medina2019robust}. In the context of FGO, m-estimators work by incorporating a robust loss function, such as the Cauchy or Huber function \cite{huber2011robust} into the standard error formulation.
Watson \textit{et.al} \cite{watson2017robust} explored robust GNSS positioning based on FGO in urban environments. They evaluate several robust optimization techniques, including m-estimators (Huber, Cauchy) and graph-based methods (switchable constraints, dynamic covariance scaling, and max-mixtures). 
Zhang \textit{et.al} \cite{zhang2024gnss} presented a continuous-time FGO framework that incorporates m-estimators, specifically the Cauchy loss function, to handle faulty GNSS observations.
Similarly, Ahmadi \textit{et.al} introduced a robust LC integration of GNSS/IMU using FGO in \cite{ahmadi2025robust}, which incorporates an m-estimator with the Huber loss function applied to GNSS factors to enhance GNSS positioning accuracy.
Ben \textit{et.al} \cite{ben2024robust} presented robust FGO method for land vehicle navigation, integrating INS, GNSS, and odometry with dynamic kernel principal component analysis. It used a dynamic m-estimator with a switchable weight function to adjust fusion weights based on fault severity, capturing nonlinear data features.
Bai \textit{et.al} introduced a GNSS positioning method, without IMU integration, using window carrier phase and FGO to enhance accuracy in urban settings in \cite{bai2022time}. It integrates pseudorange, Doppler, and carrier-phase measurements across multiple epochs. The method used the Cauchy-based m-estimator, which robustly mitigates cycle slip effects, outperforming the Huber-based m-estimator.

Despite these advancements, existing methods face several limitations. While adaptive techniques such
as map-aided fusion and virtual constraints improve resilience, they often rely on external data or predefined models, which may not fully adapt to rapidly changing environmental conditions or diverse interference scenarios. Dynamic weighting schemes, as explored in \cite{pan2024smartphone}, show promise but lack sophisticated mechanisms to continuously assess and adjust sensor reliability in real time. Furthermore, handling carrier-phase ambiguities and non-Gaussian noise in highly dynamic environments remains an area requiring further exploration. 
Moreover, the performance of the m-estimator is still limited by parameter tuning. The research in \cite{pfeifer2017dynamic, watson2017robust} shows the extensive parameter tuning is required to obtain satisfactory performance using the m-estimator.
Inspired by the aforementioned works, we address the problem 
by introducing an adaptive factor graph optimization framework featuring built-in robustness through adaptive loss modeling.
Leveraging the Barron loss within a TC GNSS/IMU fusion context, our approach eliminates the need for heuristic parameter tuning and reliance on external aiding data.
Moreover, it is a flexible loss function that generalizes many common robust loss functions like L2, L1, Cauchy, etc., into one unified formulation.
This capability is particularly critical in urban and obstructed environments.

\section{MATHEMATICAL BACKGROUND} \label{BACKGROUND}

\subsection{Frame Definitions}
The integration of multi-sensor data in this work necessitates a precise description of the coordinate systems, which are detailed in the following
\cite{groves2015principles}:
\begin{itemize}
    \item Earth-Centered Earth-Fixed (ECEF) coordinate frame: It is a global Cartesian frame with its origin at the Earth's center of mass. The $x$-axis points to the Prime Meridian, the $y$-axis lies in the equatorial plane and is orthogonal to the $x$-axis, and the $z$-axis points toward the North Pole. GNSS positions and velocities are typically represented in this frame for global referencing. 
    \item Local tangent plane (East-North-Up, ENU) coordinate frame: It is a local reference frame tangent to the Earth’s surface at a specified reference point. The $x$-axis points eastward, $y$-axis points northward, and the $z$-axis points upward, orthogonal to the tangent plane. This frame is useful for representing relative positions, motion, and GNSS errors in a localized context.
    \item Body frame: It is a coordinate frame fixed to the moving object (vehicle or UAV). The $x$-axis is aligned with the forward direction of motion, $y$-axis points to the right, and the $z$-axis points downward. IMU measurements, including accelerations and angular velocities, are expressed in this frame.
    \item Geodetic coordinate frame: The geodetic frame represents positions using latitude $\phi$, longitude $\lambda$, and altitude $h$, relative to the WGS84 ellipsoid. GNSS outputs are typically provided in this format and transformed into other coordinate systems for integration.
\end{itemize}
\begin{figure}[tb!]
    \centering
    \includegraphics[width=0.75\linewidth]{ 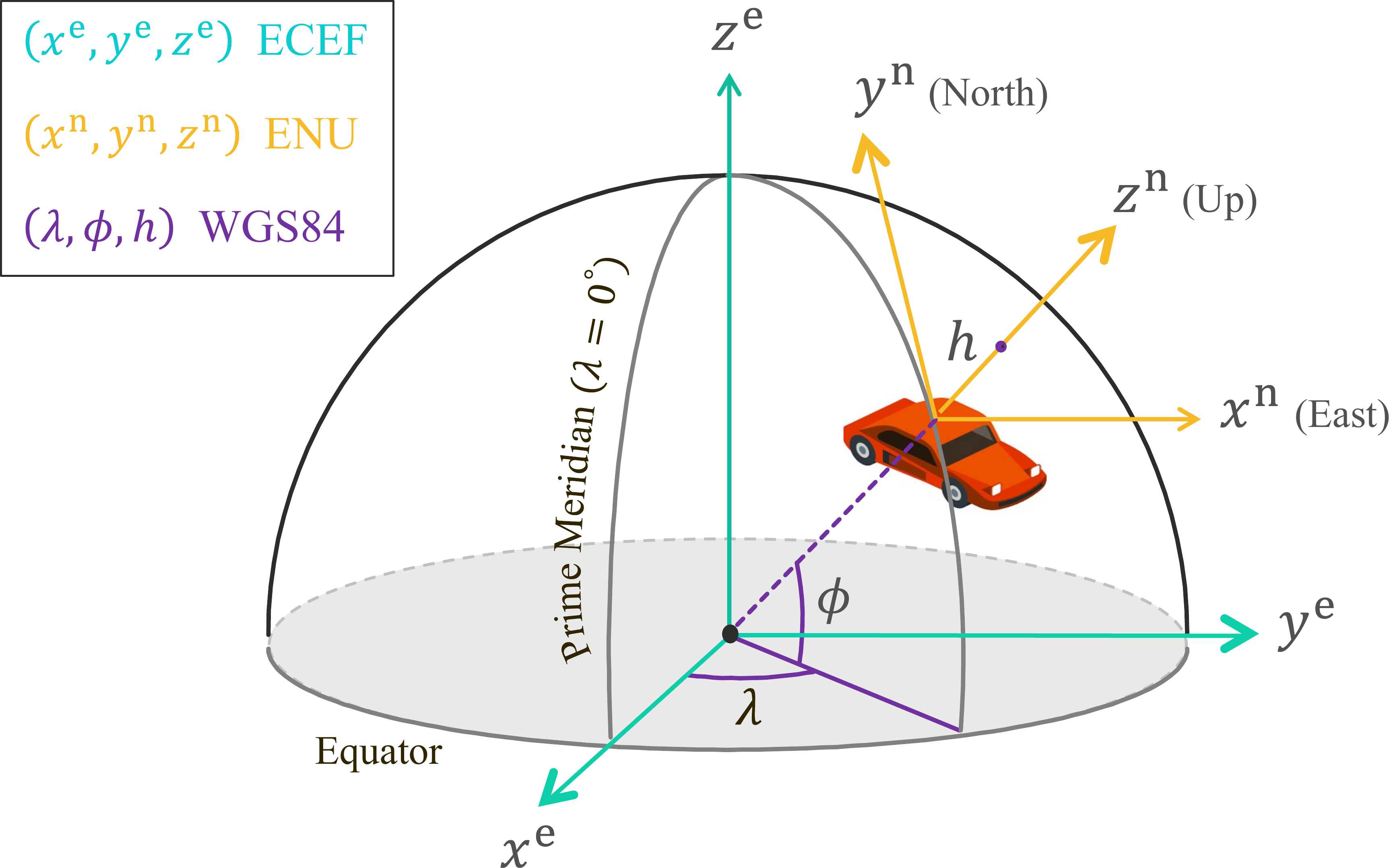}
    \caption{Illustration of the coordinate frames used in this paper; ECEF $(\cdot)^\mathrm{e}$, ENU $(\cdot)^\mathrm{n}$, and World Geodetic System (WGS84) frames.}
    \label{fig_frames}
\end{figure}

For consistent GNSS–IMU integration, transformations between coordinate systems are essential.
Figure \ref{fig_frames} depicts these systems and their relationships as the basis for the integration strategy.
GNSS positions, provided in geodetic coordinates ($\phi, \lambda, h$), are converted to ECEF Cartesian coordinates using the WGS84 ellipsoid model. These are further transformed into ENU coordinates via a rotation matrix based on the latitude and longitude of a reference point. IMU measurements, expressed in the body frame, are transformed into the ENU frame using orientation information such as quaternions or rotation matrices. Sensor data from individual devices are aligned with the body frame through extrinsic calibration parameters.

\subsection{Maximum a Posteriori Estimation}
Maximum a posteriori (MAP) estimation is a statistical inference method that provides a way to estimate the parameters of a probabilistic model given observed data, using Bayes’ rule as its foundation. It extends the concept of maximum likelihood estimation (MLE) by incorporating prior knowledge about the parameters through the use of a prior distribution.
MAP estimation seeks to find the parameter $\mathbf{x}$ that maximizes the posterior probability $p(\mathbf{x} | \mathbf{z}) \propto p(\mathbf{z} | \mathbf{x}) p(\mathbf{x})$, given observations $\mathbf{z}$:
\begin{equation}
    \mathbf{x^*} = \arg\max_{\mathbf{x}} p(\mathbf{x} | \mathbf{z}),
    \label{MAP_formula}
\end{equation}
or equivalently, by minimizing the negative log-posterior:
\begin{equation*}
\mathbf{x^*} = \arg\min_{\mathbf{x}} \big[ -\log p(\mathbf{z} | \mathbf{x}) - \log p(\mathbf{x}) \big].
\end{equation*}
This formulation enables the integration of measurements and priors into a unified optimization problem, forming the basis for techniques such as Kalman filtering and factor graph optimization.

\subsection{Robust Statistic}
Robust statistics is a branch of statistics designed to improve the performance of estimators in the presence of outliers, noise, or deviations from assumed data distributions \cite{huber2011robust}.  Unlike classical methods, which rely heavily on assumptions like Gaussian noise, robust methods aim to provide reliable estimates even when these assumptions are violated. Robust loss functions are particularly useful in machine learning, computer vision, and sensor fusion applications. 

Key concepts in robust statistics include the use of m-estimators, which generalize MLE through the incorporation of loss functions beyond the quadratic form.
While the class of m-estimators includes non-robust instances such as the least mean squares (LMS) estimator, robust m-estimators are specifically designed to mitigate the effect of outliers.
Outlier measurements can dominate the residual and lead to local minimum estimation. Instead of de-weighting the outlier measurements by tuning the covariance matrix, the robust $\zeta$-function of robust m-estimator takes care of de-weighting. 
Unlike standard least-squares, which minimize the sum of squared residuals, robust m-estimators allow for alternative loss functions that reduce the impact of outliers as:
\begin{equation}
    \min_{\mathbf{x}} \sum_{i} \zeta(r_i),
\end{equation}
where $r_i$ is the residual of the $i$-th measurement, and $\zeta(r)$ is a robust loss function. Common robust loss functions include:
\begin{itemize}
    \item Huber loss (quadratic for small residuals and linear for large residuals):
    \begin{equation} \label{Huber}
        \zeta(r) =
        \begin{cases}
            \frac{1}{2}r^2, & \text{if } |r| \leq \alpha_\mathrm{h}, \\
            \alpha_\mathrm{h} |r| - \frac{1}{2} \alpha_\mathrm{h}^2, & \text{if } |r| > \alpha_\mathrm{h},
        \end{cases}
    \end{equation}
    where $\alpha_\mathrm{h}$ is a tuning parameter that defines the threshold for treating residuals as outliers. In summary, the choice of $\alpha_\mathrm{h}$ should reflect the possible proportion of outliers in the data. Therefore, it is sensible to adjust the $\alpha_\mathrm{h}$ value accordingly based on the distribution of the data. 

    \item Tukey loss (quadratic for small residuals and constant for large ones):
    \begin{equation} \label{Tukey}
        \zeta(r) =
        \begin{cases}
            \frac{\alpha_\mathrm{t}^2}{6} \left(1 - \left(1 - \frac{r^2}{\alpha_\mathrm{t}^2}\right)^3\right), & \text{if } |r| \leq \alpha_\mathrm{t}, \\
            \frac{\alpha_\mathrm{t}^2}{6}, & \text{if } |r| > \alpha_\mathrm{t},
        \end{cases}
    \end{equation}
    where $\alpha_\mathrm{t}$ controls the range of residuals.

    \item Cauchy loss (quadratic for small residuals, logarithmic for large outliers):
    \begin{equation} \label{Cauchy}
        \zeta(r) = \frac{\alpha_\mathrm{c}^2}{2} \ln\left(1 + \frac{r^2}{\alpha_\mathrm{c}^2}\right),
    \end{equation}
    where $\alpha_\mathrm{c}$ adjusts the influence of large residuals.
\end{itemize}


However, every m-estimator’s robust loss function has one or more tuning parameters that control the influence of different data.
Selecting an appropriate m-estimator and tuning these parameters introduce a degree of uncertainty, particularly when applying the method to new datasets or new problems.
This challenge is particularly relevant in GNSS/IMU integration using FGO, where robust statistics helps mitigate the impact of GNSS outliers caused by multipath or interference, enhancing positioning accuracy and reliability in challenging conditions.

\subsection{States}
The system states to be estimated include the platform's position $\mathbf{p}$, the receiver clock bias $\delta_{\mathrm{r}, k}$, the velocity $\mathbf{v}$, and the orientation $\mathbf{q}$. The position and velocity are typically expressed in the ENU or ECEF frame, while the orientation is represented as a quaternion or a rotation matrix that transforms from the body frame to the navigation frame. Additionally, the biases of the IMU sensors, including accelerometer bias $\mathbf{b}_\mathrm{a}$ and gyroscope bias $\mathbf{b}_\mathrm{g}$ are estimated. The system state at any time step $k$ can be presented as
$
    \mathbf{x}_k = \begin{bmatrix}
        \mathbf{p}_k & \mathbf{v}_k & \mathbf{q}_k & 
        \delta_{\mathrm{r}, k} &
        \mathbf{b}_{\mathrm{a}, k} & \mathbf{b}_{\mathrm{g}, k} 
    \end{bmatrix}^\top.
$

\subsection{Notation}
Throughout this paper, lowercase bold letters denote vectors, and uppercase bold letters denote matrices. Scalars are represented by lowercase letters. The superscript $(\cdot)^\top$ denotes the transpose of a vector or matrix, while the operator $(\cdot)^\wedge$ maps a 3D vector in $\mathbb{R}^3$ to its corresponding skew symmetric matrix in the Lie algebra $\mathfrak{so}(3)$. The exponential map $\exp(\cdot)$ converts elements from $\mathfrak{so}(3)$ to the rotation group $\mathrm{SO}(3)$. Time-dependent variables are indexed using subscripts; for instance, $\mathbf{x}_k$ indicates the state vector at time step $k$. Subscripts $\mathrm{r}$, $\mathrm{s}$, and $k$ are used to represent the receiver, satellite, and time step indices, respectively.

\section{PROBLEM FORMULATION} \label{formulation}
\begin{figure}[tb!]
    \centering
    \includegraphics[width=1.03\linewidth]{ 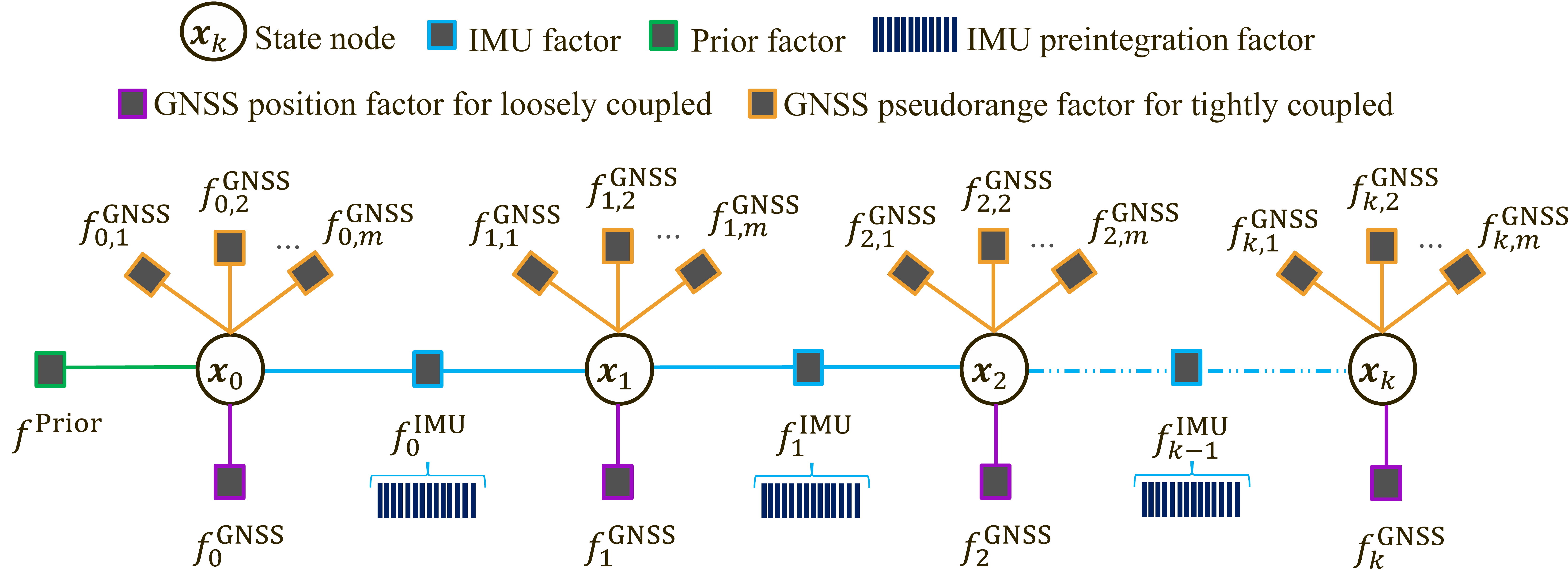}
    \caption{Graph structure of the implemented loosely coupled and tightly coupled GNSS/IMU integration based on FGO.}
    \label{fgo_structure}
\end{figure}
Fusion techniques can range from simple or weighted averaging of sensor measurements, to intermediate approaches like weighted least mean squares, and further to advanced probabilistic methods including Kalman filters, EKF, and FGO. Among these, FGO has gained increasing attention as a promising alternative to EKF for GNSS/IMU integration, offering improved robustness and accuracy \cite{wen2021factor}.
%
Its theoretical underpinnings lie in probabilistic graphical models and non-linear optimization techniques.
A factor graph is a type of probabilistic graphical model denoted as $G = (F, \mathbf{x}, E)$, with nodes $\mathbf{x}_j \in \mathbf{x}$ represent the unknown variables (position, orientation, velocity), and factors $f_i \in F$ represent constraints or measurements that relate these variables \cite{dellaert2017factor}.
Edges $e_{ij} \in E$ connect factor nodes to variable nodes only if the factor $f_i$ involves the variable $x_j$. The factor graph $G$ expresses the function $f(\mathbf{x})$ as a product of factors:
\begin{equation}
f(\mathbf{x}) = \prod_{i} f_i(\mathbf{x}_i),
\end{equation}
where $\mathbf{x}_i$ represents the subset of variables connected to the factor $f_i$ by edges in the graph.
FGO addresses the sensor fusion problem as a MAP estimation, where the entire set of states is optimized by minimizing an objective function, typically an m-estimator, such as the sum of squared residuals (errors) across all factors in the graph:
\begin{equation}
    \mathbf{x}^* = \arg \max_\mathbf{x} \left( \prod_{i} f_i(\mathbf{x}_i) \right).
\end{equation}
When the measurement noise follows a Gaussian
distribution, the negative logarithm of probability
distribution becomes proportional to the error function. This reformulates the MAP estimation problem as a nonlinear least-squares optimization problem:
\begin{equation}\label{FGO_NLS}
    \mathbf{x}^* 
    = \arg\min_{\mathbf{x}} \sum_{i, j} \| e(\mathbf{x}_i, \mathbf{x}_j, \mathbf{z}_{i,j}) \|^2_{\mathbf{\Sigma}_{i, j}}, 
\end{equation}
where $\mathbf{\Sigma}_{i,j}$ is the known covariance matrix of the measurement vector, and $\| e \|_{\mathbf{\Sigma}}^2 = e^\top \cdot \mathbf{\Sigma}^{-1} \cdot e $ denotes the squared Mahalanobis norm of $e$ with $e = \mathbf{z}_{i,j} - h(\mathbf{x}_i, \mathbf{x}_j)$ is the difference between the received measurement and the measurement expected for the true states.
The method can also be adapted to other m-estimators based on Huber, Tukey or Cauchy loss functions for example, but the objective function may be non-convex, and then the initialization matters.
Solving optimization problem \eqref{FGO_NLS} typically involves iterative nonlinear least-squares algorithms such as Gauss-Newton method \cite{wang2012gauss}
and the Levenberg-Marquardt algorithm (LMA) \cite{gavin2019levenberg}. 
When new observations are added to the factor graph, the solution can be updated incrementally using a method like incremental smoothing and mapping (iSAM2) algorithm \cite{kaess2012isam2}. This method efficiently updates only the relevant subset of the graph affected by the new observations, typically employing LMA for the local optimization, enabling real-time solutions.

Unlike traditional Bayesian filters such as EKF, the FGO captures the posterior probability of states over time, integrating both past measurements and system updates to optimize the entire state set. 
Historical information plays a crucial role in FGO,
where all measurements and states are encoded into a factor graph, and the sensor fusion problem is iteratively tackled through non-linear optimization methods. Consequently, errors arising from linearization steps are minimized.
Figure \ref{fgo_structure} shows the graph structure employed in this work for GNSS/IMU integration, demonstrating both loosely and tightly coupled configurations within the FGO-based framework.

\subsection{GNSS Position Factor}
The GNSS factor serves as a key component to provide position and velocity measurements obtained from the GNSS receiver. They are formulated as a residual term that relates the estimated states to the position and/or velocity measurements provided by the GNSS. For a given state $\mathbf{x}_k$ at time $k$, the GNSS factor is defined as:
\begin{equation}
    \mathbf{z}_k^{\mathrm{GNSS}} = \mathbf{h}^{\mathrm{GNSS}} (\mathbf{x}_k) + \mathbf{\omega}^\mathrm{GNSS},
    \label{GNSS_factor}
\end{equation}
where $\mathbf{h}^{\mathrm{GNSS}} (\mathbf{x}_k)$ is the measurement model that maps the state $\mathbf{x}_k$ to the measurement space and  $\mathbf{\omega}^\mathrm{GNSS}$ is the noise of GNSS measurements.
The GNSS factor residual is typically weighted by the measurement covariance $\mathbf{\Sigma}^\mathrm{GNSS}_k$, resulting in the following contribution to the objective function:
\begin{equation}
    \| \mathbf{e}_k^{\mathrm{GNSS}} \|^2_{\mathbf{\Sigma}_k^\mathrm{GNSS}} = \| \mathbf{z}_k^{\mathrm{GNSS}} - \mathbf{h}^{\mathrm{GNSS}} (\mathbf{x}_k) \|^2_{\mathbf{\Sigma}_k^\mathrm{GNSS}}.
    \label{GNSS_error}
\end{equation}

\subsection{GNSS Pseudorange Factor}
Pseudorange measurements play a crucial role in GNSS positioning algorithms and are a key input for sensor fusion techniques. In tightly coupled integration, pseudoranges provide essential position information that can be fused with inertial measurements to improve positioning accuracy and robustness, especially in environments with degraded GNSS signals.
A pseudorange $\rho$ is a measurement of the signal propagation distance between a GNSS satellite and a receiver. It is referred to as "pseudo" because it is subject to various errors and biases, including satellite and receiver clock offsets, atmospheric delays, multipath effects, and measurement noise. Consequently, the pseudorange must be corrected to accurately reflect the true range between the satellite and the receiver. The pseudorange measurement equation can be written as:
\begin{equation} \label{Pseudorange_equation}
    \rho^\mathrm{s}_{\mathrm{r}, k} = \mathbf{p}^\mathrm{s}_{\mathrm{r}, k} + c_\mathrm{l} \left ( \delta_{\mathrm{r}, k} - \delta^\mathrm{s}_{\mathrm{r}, k} \right ) + \mathbf{I}^\mathrm{s}_{\mathrm{r}, k} + \mathbf{T}^\mathrm{s}_{\mathrm{r}, k} + \varepsilon^\mathrm{s}_{\mathrm{r}, k},
\end{equation}
where indices $\mathrm{s}$ and $\mathrm{r}$ denote satellite and receiver, respectively, $c_\mathrm{l}$ is the speed of the light, $\delta_{\mathrm{r}, k}$ is the receiver clock bias, $\delta^\mathrm{s}_{\mathrm{r}, k}$ is the satellite clock bias, $\mathbf{I}^\mathrm{s}_{\mathrm{r}, k}$ is the Ionospheric delay, $\mathbf{T}^\mathrm{s}_{\mathrm{r}, k}$ is Tropospheric delay, $\varepsilon^\mathrm{s}_{\mathrm{r}, k}$ can show multipath effect, NLOS receptions, receiver noise, etc., and $\mathbf{p}^\mathrm{s}_{\mathrm{r}, k}$ is range distance given by:
\begin{equation} \label{range_equation}
    \mathbf{p}^\mathrm{s}_{\mathrm{r}, k} = \sqrt{
    \left( x_k^\mathrm{s} - x_{\mathrm{r}, k} \right)^2 +
    \left( y_k^\mathrm{s} - y_{\mathrm{r}, k} \right)^2 +
    \left( z_k^\mathrm{s} - z_{\mathrm{r}, k} \right)^2
    },
\end{equation}
where $\mathbf{p}_k^\mathrm{s} = \begin{bmatrix}
x_k^\mathrm{s} & y_k^\mathrm{s} & z_k^\mathrm{s} 
\end{bmatrix}^\top$ is the satellite position, 
$\mathbf{p}_{\mathrm{r}, k} = \begin{bmatrix} x_{\mathrm{r}, k} & y_{\mathrm{r}, k} & z_{\mathrm{r}, k} \end{bmatrix}^\top$ is the receiver position.
The measurement model for a GNSS pseudorange observation is given by:
\begin{equation}
    \rho^\mathrm{s}_{\mathrm{r}, k} = \mathbf{h}^{\rho} (\mathbf{p}_k^s, \mathbf{p}_{\mathrm{r}, k}, \delta_{\mathrm{r}, k}) + \mathbf{\omega}^\rho,
\end{equation}
where the measurement function is defined as $\mathbf{h}^{\rho} (\cdot) = \| \mathbf{p}_k^s - \mathbf{p}_{\mathrm{r}, k} \| + \delta_{\mathrm{r}, k}$, and $\mathbf{\omega}^\rho$ denotes the noise of pseudorange measurement. 
The corresponding error function for a given satellite measurement is:
\begin{equation}
    \| \mathbf{e}_k^{\rho} \|^2_{\mathbf{\Sigma}_k^\rho} = \| \rho^\mathrm{s}_{\mathrm{r}, k} - \mathbf{h}^{\rho} (\mathbf{p}_k^s, \mathbf{p}_{\mathrm{r}, k}, \delta_{\mathrm{r}, k}) \|^2_{\mathbf{\Sigma}_k^\rho},
    \label{pseudorange_error}
\end{equation}
where $\mathbf{\Sigma}_k^\rho$ is the associated  covariance matrix. The unknown variables to be estimated are the receiver position components $x_{\mathrm{r}, k}, \, y_{\mathrm{r}, \, k}, z_{\mathrm{r}, k}$, and the clock bias $\delta_{\mathrm{r}, k}$.


\subsection{IMU Factor}
The IMU factor serves as a key component in providing motion updates, including acceleration and angular velocity.
IMU outputs have a higher frequency and better short-term positioning accuracy compared to GNSS measurements. To meet real-time navigation requirements, the IMU pre-integration technique introduced in \cite{lupton2011visual, forster2016manifold} is used.
IMU measurements are integrated between GNSS updates as factors between states, representing relative motion constraints in the graph (see Fig.~\ref{fgo_structure}). It can help to bridge the gaps when GNSS data is unavailable or unreliable. We use this IMU mechanism to formulate the IMU factor.

Given two consecutive states $\mathbf{x}_k$ and $\mathbf{x}_{k+1}$, the nonlinear measurement equation is defined as:
\begin{equation}
    \mathbf{x}_{k+1} = \mathbf{h}^{\mathrm{IMU}} (\mathbf{x}_k, \mathbf{u}_k) + \mathbf{\omega}^\mathrm{IMU},
    \label{IMU_factor}
\end{equation}
where $\mathbf{h}^{\mathrm{IMU}} (\mathbf{x}_k, \mathbf{u}_k)$ is the IMU-based motion model that predicts the next state using the current state $\mathbf{x}_k$ and the IMU measurements $\mathbf{u}_k$, and  $\mathbf{\omega}^\mathrm{IMU}$ is the IMU measurement noise.
The error function for the IMU factor weighted by the covariance matrix $\mathbf{\Sigma}^\mathrm{IMU}$ is defined as:
\begin{equation}
    \| \mathbf{e}_k^{\mathrm{IMU}} \|^2_{\mathbf{\Sigma}_k^\mathrm{IMU}} = \| \mathbf{x}_{k+1} - \mathbf{h}^{\mathrm{IMU}} (\mathbf{x}_k, \mathbf{u}_k) \|^2_{\mathbf{\Sigma}_k^\mathrm{IMU}}.
    \label{IMU_error}
\end{equation}

\subsubsection{IMU Pre-integration} \label{imu_pre_integration_section}
IMU measurements typically arrive at a much higher frequency (e.g., $100-1000$ Hz) than GNSS measurements (e.g., $1-10$ Hz). Creating a factor graph node for every IMU measurement would lead to an intractably large graph. IMU preintegration addresses this by analytically integrating the IMU measurements between two consecutive keyframes to form a single relative motion constraint (factor) that summarizes the IMU information over that interval.

The pioneering work of Forster et al. \cite{forster2016manifold} laid the foundation for IMU preintegration, particularly addressing the non-Euclidean nature of rotations by performing integration on the SO(3) manifold.
The key advantage of this approach is that the subsequent updates to the keyframe states (poses, velocities) during optimization do not require re-integration. 
The IMU kinematics describe the evolution of rotation $\mathbf{R}$, velocity $\mathbf{v}$, and position $\mathbf{p}$:
\begin{equation}
    \begin{aligned}
    \dot{\mathbf{R}}_\mathrm{WB}(t) &= \mathbf{R}_\mathrm{WB}(t)\left( \tilde{\bm{\omega}}(t) - \mathbf{b}_\mathrm{g}(t) - \bm{\eta}_g^d(t) \right)^\wedge \\
    \dot{\mathbf{v}}^\mathrm{W}(t) &= \mathbf{R}_\mathrm{WB}(t)\left( \tilde{\mathbf{a}}(t) - \mathbf{b}_\mathrm{a}(t) - \bm{\eta}_a^d(t) \right) + \mathbf{g} \\
    \dot{\mathbf{p}}^\mathrm{W}(t) &= \mathbf{v}^\mathrm{W}(t)
    \end{aligned}
\end{equation}
where $\mathbf{R}_\mathrm{WB}$ is the rotation from body frame to world frame, $\tilde{\bm{\omega}}$ and $\tilde{\mathbf{a}}$ are the raw gyroscope and accelerometer measurements, $\mathbf{b}_\mathrm{g}$ and $\mathbf{b}_\mathrm{a}$ are their respective biases, $\bm{\eta}_g^d$ and $\bm{\eta}_a^d$ are zero-mean Gaussian noise terms, and $\mathbf{g}$ is the gravity vector.

Between two keyframes at times $t_i$ and $t_j$, the preintegrated measurements define the relative change in orientation $\Delta \mathbf{R}_{ij}$, velocity $\Delta \mathbf{v}_{ij}$, and position $\Delta \mathbf{p}_{ij}$, expressed in the local frame of keyframe $i$:
\begin{equation}\label{pre_2}
    \begin{aligned}
    \Delta \mathbf{R}_{ij} &= \mathbf{R}_i^\top \mathbf{R}_j \\
    \Delta \mathbf{v}_{ij} &= \mathbf{R}_i^\top (\mathbf{v}_j - \mathbf{v}_i - \mathbf{g} \Delta t_{ij}) \\
    \Delta \mathbf{p}_{ij} &= \mathbf{R}_i^\top (\mathbf{p}_j - \mathbf{p}_i - \mathbf{v}_i \Delta t_{ij} - \tfrac{1}{2} \mathbf{g} \Delta t_{ij}^2)
    \end{aligned}
\end{equation}
where $\Delta t_{ij} = t_j - t_i$.
These $\Delta$ terms are obtained by integrating the IMU measurements over the interval $[t_i, t_j]$, assuming the biases $\mathbf{b}_\mathrm{g}$ and $\mathbf{b}_\mathrm{a}$ are constant within this short interval but can change between intervals:
\begin{equation} \label{pre_3}
    \begin{aligned}
    \Delta \tilde{\mathbf{R}}_{ij}(\mathbf{b}_g) &= \prod_{k=i}^{j-1} \exp\left( (\tilde{\bm{\omega}}_k - \mathbf{b}_g) \Delta t_k \right) \\
    \Delta \tilde{\mathbf{v}}_{ij}(\mathbf{b}_g, \mathbf{b}_a) &= \sum_{k=i}^{j-1} \Delta \tilde{\mathbf{R}}_{ik}(\mathbf{b}_g) \left( \tilde{\mathbf{a}}_k - \mathbf{b}_a \right) \Delta t_k \\
    \Delta \tilde{\mathbf{p}}_{ij}(\mathbf{b}_g, \mathbf{b}_a) &= \sum_{k=i}^{j-1} \frac{3}{2} \Delta \tilde{\mathbf{R}}_{ik}(\mathbf{b}_g) \left( \tilde{\mathbf{a}}_k - \mathbf{b}_a \right) \Delta t_k^2
    \end{aligned}
\end{equation}

Here, $\exp(\cdot)$ is the exponential map from $\mathfrak{so}(3)$ to $\mathrm{SO}(3)$. These $\Delta \tilde{\mathbf{R}}_{ij}$ (preintegrated rotation
measurement), $\Delta \tilde{\mathbf{v}}_{ij}$ (preintegrated velocity measurement), and $\Delta \tilde{\mathbf{p}}_{ij}$ (preintegrated position measurement) are the actual measurements provided by the IMU preintegration.
See \cite{forster2016manifold} for more details.
The error function for the IMU factor then compares these preintegrated measurements with the values predicted from the estimated states.

\subsection{Nonlinear Least-Squares Optimization} \label{sec_FGO_NLS}
\begin{figure}[tb!]
    \centering
    \includegraphics[width=\linewidth]{ 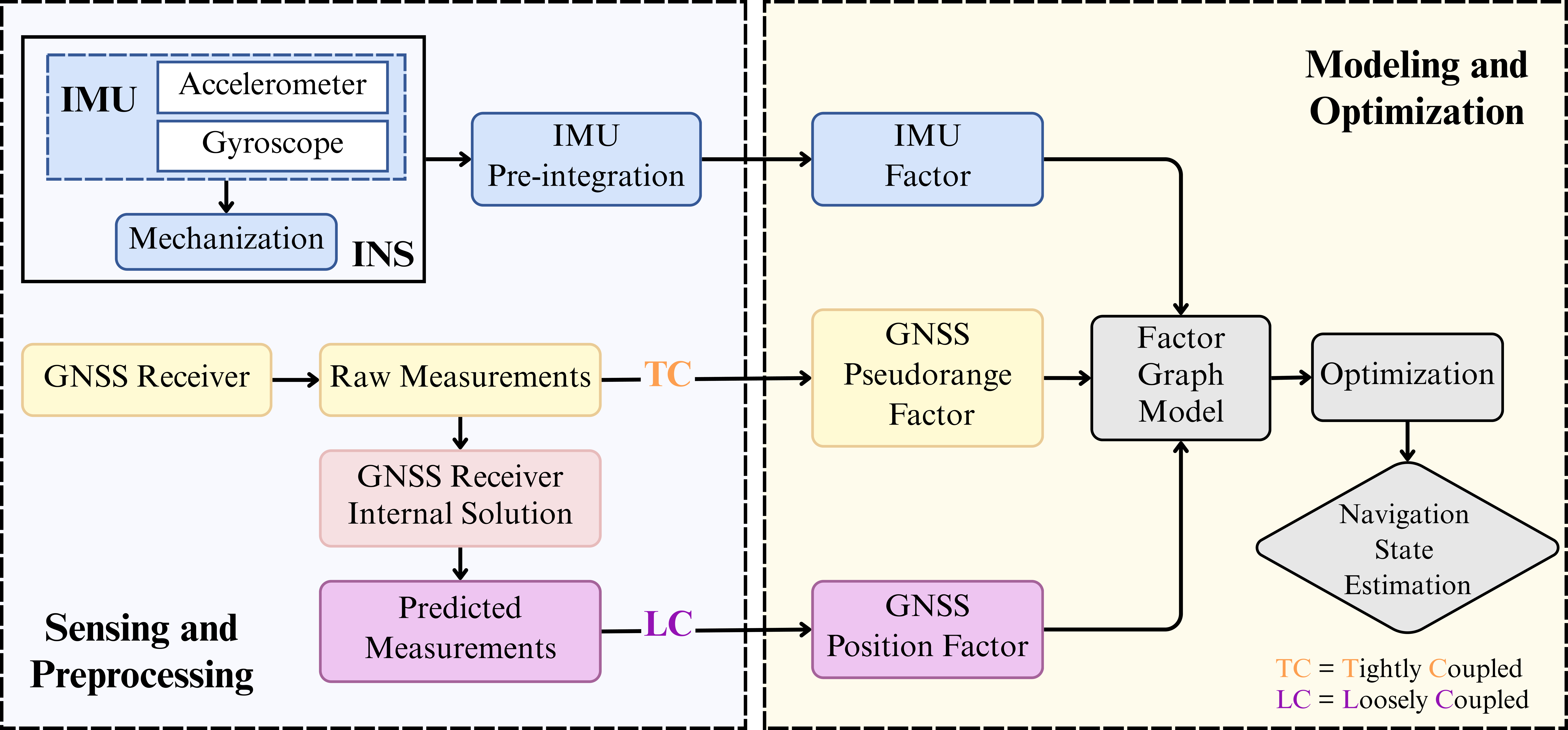}
    \caption{Flowchart of the implemented loosely coupled/tightly coupled GNSS/IMU integration based on FGO.}
    \label{fgo_flowchart}
\end{figure}

Building on the factors derived above, we implement a LC and TC fusion of GNSS and IMU measurements
, where LC typically underperforms TC methods \cite{wen2021factor}.
The overall system flow is depicted in Fig.~\ref{fgo_flowchart}.

\subsubsection{FGO-based LC integration}
In LC approaches, processed GNSS position and velocity estimates are integrated with other sensors, like an IMU, offering ease of implementation in both hardware and software.
Based on the two factor types considered in this work, the optimal state vector $\mathbf{x}^*$ is computed by solving the following optimization problem:
\begin{equation}
    \mathbf{x}_{\mathrm{LC}}^* = \arg \min_\mathbf{x} \sum_k \| \mathbf{e}_k^{\mathrm{GNSS}} \|^2_{\mathbf{\Sigma}_k^\mathrm{GNSS}} + \sum_k \| \mathbf{e}_k^{\mathrm{IMU}} \|^2_{\mathbf{\Sigma}_k^\mathrm{IMU}}.
    \label{LC_GNSS_IMU_eq}
\end{equation}

\subsubsection{FGO-based TC integration}
Unlike LC approaches, TC methods fuse raw GNSS pseudoranges with IMU measurements, providing additional constraints from multiple observed satellites to the state variables. The optimization problem is formulated as:
\begin{equation}
    \mathbf{x}_{\mathrm{TC}}^* = \arg \min_\mathbf{x} \sum_k \| \mathbf{e}_k^{\rho} \|^2_{\mathbf{\Sigma}_k^\rho} + \sum_k \| \mathbf{e}_k^{\mathrm{IMU}} \|^2_{\mathbf{\Sigma}_k^\mathrm{IMU}}.
    \label{TC_GNSS_IMU_eq}
\end{equation}

\subsection{Adaptive and Robust FGO}
A general and adaptive framework for robust loss functions was recently introduced by Barron \cite{barron2019general}, which generalizes several popular m-estimator robust loss functions by defining a single shape parameter. Additionally, the framework supports adaptive tuning of its parameters during optimization.
The basic form of the general robust loss function is given by:
\begin{equation} \label{general_robust|_loss}
    f(r, \alpha, c) = \frac{|\alpha - 2|}{\alpha} \left( 
    \left( \frac{(r/c)^2}{|\alpha - 2|} + 1 \right)^{\alpha/2} - 1
    \right),
\end{equation}
where $\alpha \in \mathbb{R}$ is a shape parameter that controls the robustness of the loss, and $c > 0$ is a scale parameter that controls the size of the quadratic bowl of loss near $r=0$. This formulation is a generalization of the $L_2$ ($\alpha \rightarrow 2$), Cauchy loss ($\alpha \rightarrow 0$), Welsch loss ($\alpha \rightarrow -\infty$), Geman-McClure loss ($\alpha \rightarrow -2$), and others. This generalization enables the loss to handle varying noise distributions without redesigning the estimator. A key observation is that the loss function’s robustness to large error values increases as $\alpha$ is changed from 2 towards $-\infty$.  In this work, we employ this loss function for the first time into the formulation of a robust FGO within a TC framework, as detailed below.

\begin{figure}[tb!] 
    \centering
    \includegraphics[width=0.75\linewidth]{ 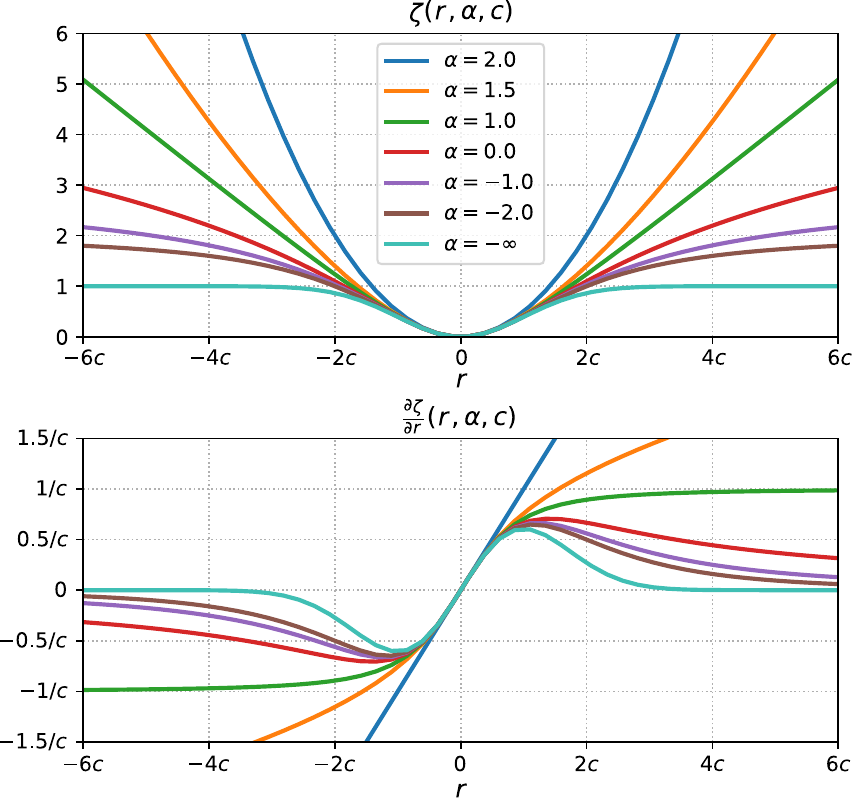}
    \caption{The general Barron loss function (top) and its gradient (bottom) for different values of its shape parameter $\alpha$.}
    \label{fig_barron_loss}
\end{figure}

The corresponding adaptive loss function $\zeta(r, \alpha, c)$ can be expressed as:
\begin{equation} \label{adaptive_loss}
    \zeta(r,\alpha,c)=
    \begin{cases}
   \frac{1}{2}\left(\frac{r}{c}\right)^2, & \text{if } \alpha = 2, \\
    \log\!\left(\frac{1}{2}\left(\frac{r}{c}\right)^2 + 1\right), & \text{if } \alpha = 0, \\
    1 - \exp\!\left(-\frac{1}{2}\left(\frac{r}{c}\right)^2\right), & \text{if } \alpha = -\infty, \\
    \frac{|\alpha-2|}{\alpha} \left(\left(\frac{(r/c)^2}{|\alpha-2|} + 1 \right) ^{\alpha/2} - 1 \right), & \text{otherwise.}
    \end{cases}
\end{equation}
Its derivative with respect to $r$, necessary for gradient-based optimization, is expressed as:
\begin{equation} \label{derivative_adaptive_loss}
    \frac{\partial \zeta}{\partial r}(r,\alpha,c)=
    \begin{cases}
    \frac{r}{c^2}, & \text{if } \alpha = 2, \\
    \frac{2r}{r^2 + 2c^2}, & \text{if } \alpha = 0, \\
    \frac{r}{c^2} \exp\!\left(-\frac{1}{2}\left(\frac{r}{c}\right)^2\right), & \text{if } \alpha = -\infty, \\
    \frac{r}{c^2} \left(\frac{(r/c)^2}{|\alpha-2|} + 1 \right) ^{\alpha/2 - 1}, & \text{otherwise.}
    \end{cases}
\end{equation}
Figure \ref{fig_barron_loss} illustrates the general robust loss function and its gradient for different values of the shape parameter $\alpha$, highlighting how varying $\alpha$ influences both the loss and its derivative.

FGO has been applied in LC and TC GNSS/IMU fusion with standard least-squares loss function, which we refer to as standard FGO (SFGO), in subsection \ref{sec_FGO_NLS}.
In order to enhance its robustness and also mitigate the impact of erroneous GNSS measurements, in this subsection, we propose to integrate the Barron loss function within the FGO framework, termed RFGO.
This general and adaptive robust loss function in graph optimization as defined in \eqref{adaptive_loss}, adaptively scales measurement residuals, allowing the optimization process to retain informative GNSS measurements while down-weighting outliers caused by multipath effects, NLOS signals, or interference. 

In the RFGO framework, we modify the standard least-squares loss by replacing the squared residual norm of either the GNSS factor (\(\mathbf{e}_k^{\mathrm{GNSS}}\)) for LC integration or the GNSS pseudorange factor (\(\mathbf{e}_k^{\rho}\)) for TC integration with its robust counterpart. This is expressed compactly using a unified residual \(\mathbf{e}_k^{\mathrm{GNSS}/\rho}\), which denotes either factor depending on which model is employed as:
\begin{equation}
    \mathbf{x}^* = \arg \min_\mathbf{x} \sum_k  \zeta \left(  \mathbf{e}_k^{\mathrm{GNSS}/\rho}  \right) + \sum_k \| \mathbf{e}_k^{\mathrm{IMU}} \|^2_{\mathbf{\Sigma}_k^\mathrm{IMU}},
    \label{robust_GNSS_IMU_eq}
\end{equation}
where $\zeta(\cdot)$ is the Barron loss function defined in \eqref{adaptive_loss}.
This formulation ensures a unified and continuous formulation across a broad spectrum of robust estimators. This allows a smooth transition between different robustness levels without switching between discrete loss types or re-deriving their gradients. This makes it possible to adaptively tune the estimator to the observed noise characteristics, ranging from nearly Gaussian to heavily contaminated, within a single consistent mathematical framework. Compared to traditional robust loss functions such as Huber, Tukey, or Cauchy, which have fixed behavior and limited flexibility, the Barron loss offers a powerful trade-off between robustness and statistical efficiency. This is especially beneficial in GNSS/IMU fusion, where the measurement noise distribution may vary significantly over time or across environments.

By applying the Barron loss functions into the FGO factors, our method can achieve more accurate and robust positioning even when GNSS data is noisy or unreliable. 
To implement this functionality, we extended the open-source GTSAM library to include Barron’s general loss as a new robust kernel since GTSAM’s built-in robust loss implementations do not support this loss function.

\section{EXPERIMENTAL Results}  \label{Results}
All experiments were conducted on a system featuring an Intel Core i7-13700H CPU and 32 GB of RAM. The factor graph formulation was implemented using the open-source Georgia Tech Smoothing and Mapping (GTSAM) library \cite{gtsam2022}, which offers a flexible and efficient framework for nonlinear optimization and probabilistic inference. The trajectory estimation results were optimized using the incremental smoothing and mapping algorithm, iSAM2 \cite{kaess2012isam2}. Furthermore, IMU preintegration was performed following the manifold-based approach described in \cite{forster2016manifold}, as detailed in Section~\ref{imu_pre_integration_section}.

\subsection{Datasetes used} 
We use the commonly used autonomous driving dataset UrbanNav \cite{hsu2021urbannav} for the process of simulation, verification, testing, and validation of the proposed approach. 
The UrbanNav dataset, collected in Hong Kong, China and Tokyo, Japan, is a comprehensive multi sensory dataset in dense urban environments for benchmarking positioning algorithms. It consists of environmental elements such as wide and narrow streets, medium and tall height buildings, tunnel, and one-sided buildings. The dataset includes data from GNSS receivers, IMU, wheel odometer, LiDAR, and multiple cameras.
For this study, we specifically focus on the GNSS and IMU sensor data.

\begin{figure}[tb!] 
    \centering
    \includegraphics[width=0.8\linewidth]{ 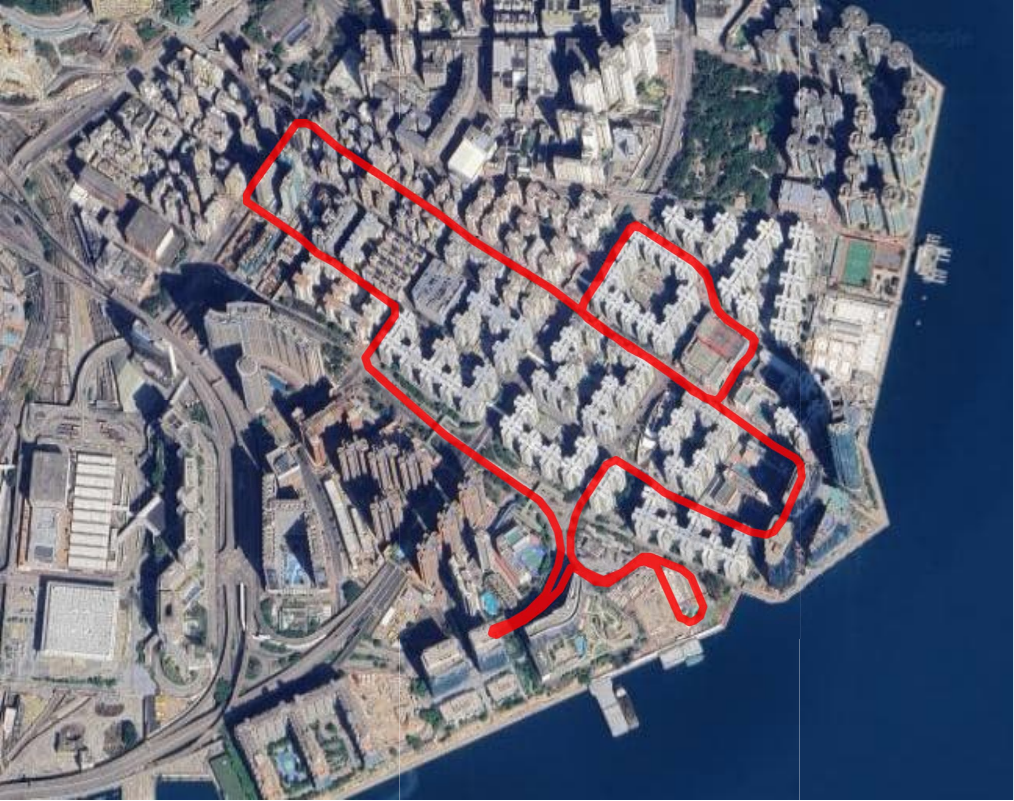}
    \caption{Overview of the UrbanNav trajectory, the deep urban canyon part of the data set, displayed on Google Maps.}.
    \label{fig_sat_urbanNav_deep}
\end{figure}

\begin{figure}[tb!] 
    \centering
    \includegraphics[width=0.85\linewidth]{ 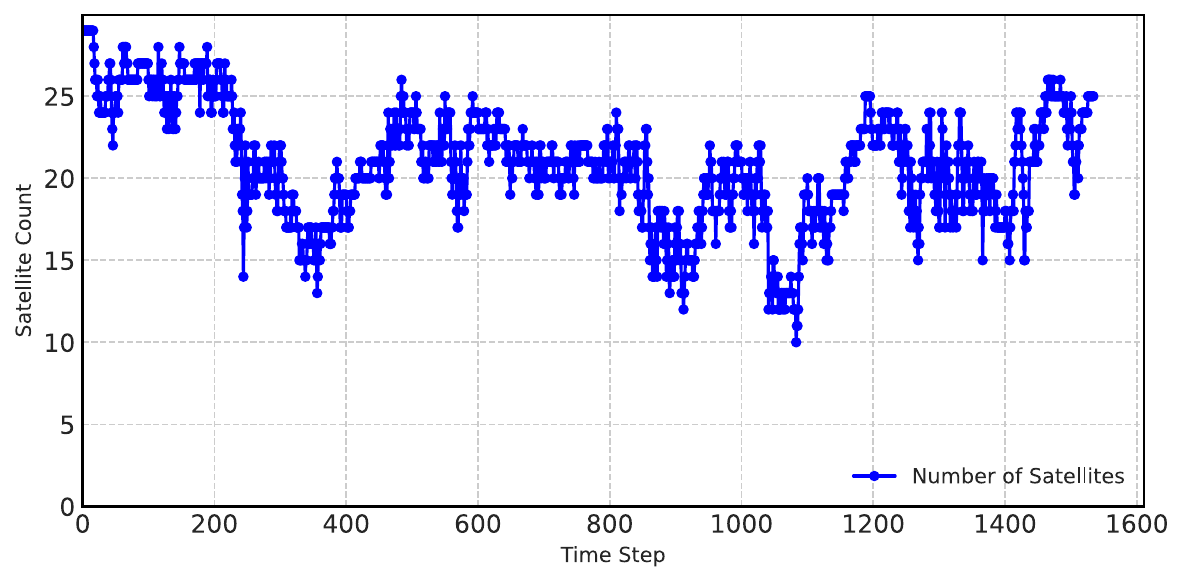}
    \caption{Number of visible GNSS satellites over time in the UrbanNav deep dataset.}
    \label{fig_num_sat}
\end{figure}

We employed the deep part of the UrbanNav dataset for evaluation, which represents a realistic and challenging GNSS-denied environment \cite{hsu2021urbannav}. 
The top view of this dataset on Google Maps with the ground truth (GT) trajectory aligned with red line is depicted in Fig.~\ref{fig_sat_urbanNav_deep}.
The trajectory begins in an open-sky region, proceeds through a wide urban street, and then transitions into a narrow street where approximately 70$\%$ of the sky is obstructed by surrounding buildings, degrading GNSS performance. The route continues through a residential area with medium- and high-rise buildings and concludes in an open-sky area, enabling loop closure. The dataset spans 1536 sec and covers a total distance of 4.5 km, encompassing a wide range of urban conditions and dynamic objects. 
As illustrated in Fig.~\ref{fig_num_sat}, the number of visible GNSS satellites varies along the route, fluctuating between 13 and 28. Sharp declines in satellite count around time steps 300, 900, and 1100 correspond to occlusions caused by urban structures, resulting in degraded GNSS observability. These fluctuations highlight the necessity for robust fusion techniques.

\subsection{RFGO Implementation Details: Parameter Tuning}
To tune the Barron loss parameters, including the shape parameter $\alpha$ and the scaling parameter $c$, we employ a data-driven hyperparameter tuning strategy informed by the residual statistics of GNSS pseudorange measurements. Specifically, we adopt a grid search over the two-dimensional parameter space $(\alpha, c)$ to identify configurations that minimize the residual-based validation error. The parameter $\alpha$ is bounded within the interval $[\alpha_\mathrm{main}, \alpha_\mathrm{max}]=[-4, 4]$, a range that generalizes a family of robust estimators, including Cauchy, Geman-McClure, Huber, and L2/MSE. The scaling parameter $c$, which controls the influence of large residuals, is constrained to $(0, c_\mathrm{max}]$, with $c_\mathrm{max} = 2$. Values of $c$ beyond 2 were found to offer no significant improvement in validation performance and may degrade robustness.
The grid consists of uniformly spaced values: $\alpha \in \{-4, -3.5, \dots, 4\}$ and $c \in \{0.1, 0.2, \dots, 2\}$. For each pair, the full graph optimization process is executed, and the performance is evaluated using the mean squared residual error over GNSS pseudorange factors.

This procedure enables adaptive selection of the loss function according to the statistical characteristics of the measurement residuals. This approach allows the resulting estimator to adapt its robustness level to the observed noise, improving generalization and resilience in GNSS-degraded environments. The final selected parameters, $\alpha = -0.75$ and $c = 1.2$, offer a loss profile suitable for mitigating GNSS outliers caused by urban occlusion and/or interferences. 
It should be noted that, in contrast to the joint optimization of $\alpha$ proposed in \cite{barron2019general}, which avoids manual tuning but requires additional regularization or explicit constraints to prevent $\alpha$ from trivially minimizing to $-\infty$, our approach determines $\alpha$ through validation based on residual statistics. This avoids the need for constraint handling during inference while still leveraging the generality and robustness of the Barron loss formulation.

\subsection{Trajectory Comparison and Performance Metrics}
To evaluate the proposed adaptive RFGO, we compare the estimated trajectories using different methods:
\begin{itemize}
    \item WLS: this method is based on the open-source RTKLIB library
    \cite{takasu2009development}. It serves as a classical GNSS-only solution, providing a
    baseline for comparison.
    \item EKF: it is implemented using TC GNSS/INS integration techniques \cite{wen2021factor}.
    \item SFGO: the TC factor graph optimization approach with standard least-squares loss function \cite{wen2021factor}.
    \item RFGO: the robust TC factor graph optimization method with general and adaptive Barron loss function.
\end{itemize}

\begin{figure}[tb!] 
    \centering
    \includegraphics[width=0.85\linewidth]{ 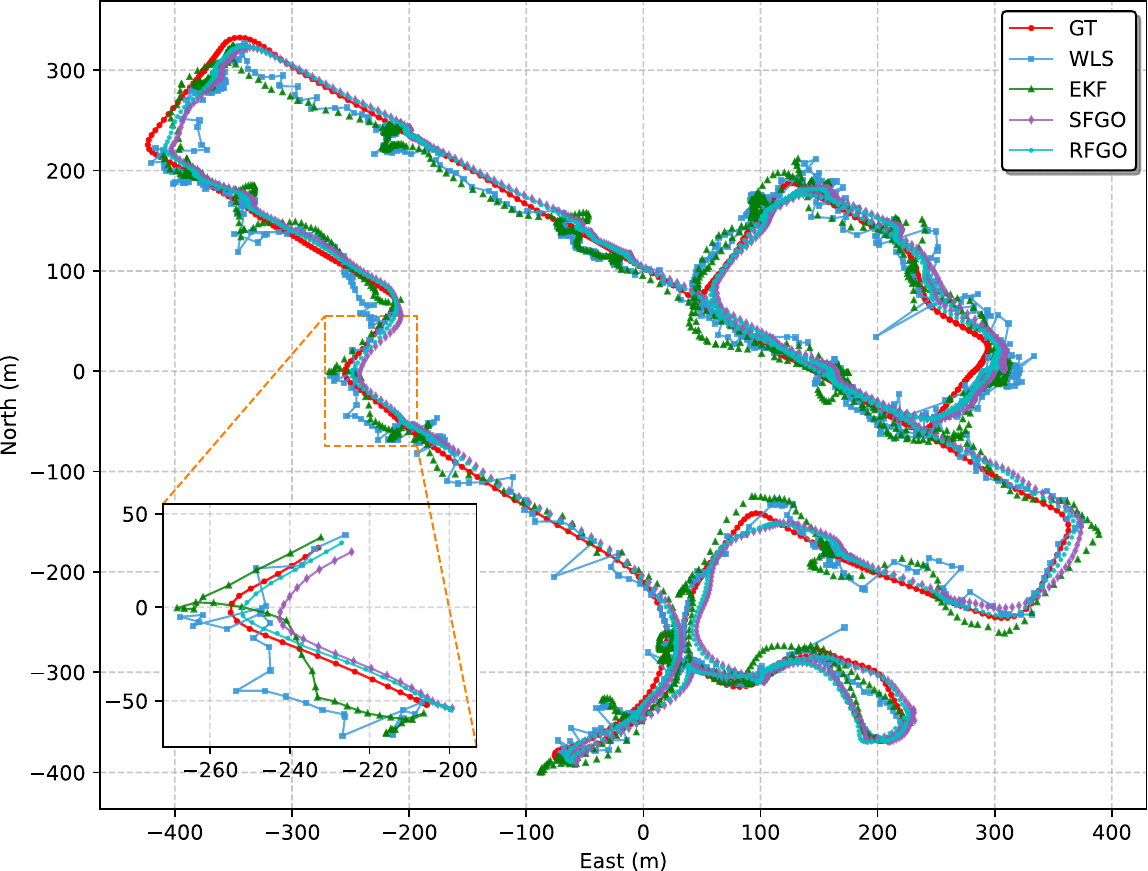}
    \caption{Estimated trajectories from different fusion methods, WLS, EKF, SFGO, and the proposed RFGO compared to the GT.}
    \label{fig_est_trajectories_rfgo}
\end{figure}

Figure \ref{fig_est_trajectories_rfgo} presents the estimated trajectories produced by various fusion methods overlaid on the GT. The WLS solution (blue) demonstrates considerable deviations, particularly in areas of degraded GNSS observability. The EKF (green) exhibits improved performance but still suffers from noticeable drift and inconsistency, especially during turns and in obscured regions. SFGO (purple) reduces the overall trajectory error compared to WLS and EKF, benefiting from the optimization-based structure of factor graphs. However, deviations remain in portions of the trajectory where GNSS measurements are heavily corrupted. In contrast, the proposed RFGO method (cyan) closely follows the GT throughout the entire trajectory, including in the zoomed-in urban canyon segment, where traditional methods diverge significantly. The improved alignment in RFGO is attributed to the adaptive robustness of the Barron loss, which dynamically attenuates the influence of outlier GNSS pseudorange measurements.
To assess the positioning accuracy of different methods, Table~\ref{tab:error_comparison} summarizes key positioning error metrics: root mean square error (RMSE), mean error (ME), maximum error (MaxE), and standard deviation (SD). 
Compared with WLS, EKF, and SFGO, RFGO achieves the lowest error across all metrics, with an RMSE of 8.13 m, ME of 6.65 m, MaxE of 18.73 m, and SD of 3.04 m. These results indicate that RFGO reduces both the overall error magnitude and variability by nearly half relative to SFGO and achieves even larger improvements compared with EKF and WLS, highlighting the effectiveness of integrating the adaptive Barron loss into the tightly coupled GNSS/IMU FGO framework.
\begin{table}[tb!]
    \centering
    \caption{Comparison of Error Metrics for Different Methods on the UrbanNav Dataset}
    \label{tab:error_comparison}
    \renewcommand{\arraystretch}{1.2} 
    \setlength{\tabcolsep}{5pt} 
    \begin{tabular}{l|c|c|c|c}
        \toprule
        \rowcolor[HTML]{FFF2CC}  
        \textbf{Method} & \textbf{RMSE}$^{\mathrm{a}}$ & \textbf{ME}$^{\mathrm{b}}$ & \textbf{MaxE}$^{\mathrm{c}}$ & \textbf{SD}$^{\mathrm{d}}$ \\
        \midrule
        WLS   & 19.67 & 16.91 & 95.43 & 10.06 \\
        EKF   & 16.33 & 14.31 & 45.44 & 7.87  \\
        SFGO  & 13.79 & 12.69 & 44.74 & 5.39  \\
        RFGO  & 8.13 & 6.65  & 18.73 & 3.04  \\
        \bottomrule
        \multicolumn{2}{l}{$^{\mathrm{a}}$ Root Mean Squared Error (RMSE).} &
        \multicolumn{3}{l}{$^{\mathrm{b}}$ Mean Error (ME).} \\
        \multicolumn{2}{l}{$^{\mathrm{c}}$ Max Error (MaxE).} &
        \multicolumn{3}{l}{$^{\mathrm{d}}$ Standard Deviation (SD).} \\
    \end{tabular}
\end{table}

\begin{figure}[tb!] 
    \centering
    \includegraphics[width=0.85\linewidth]{ 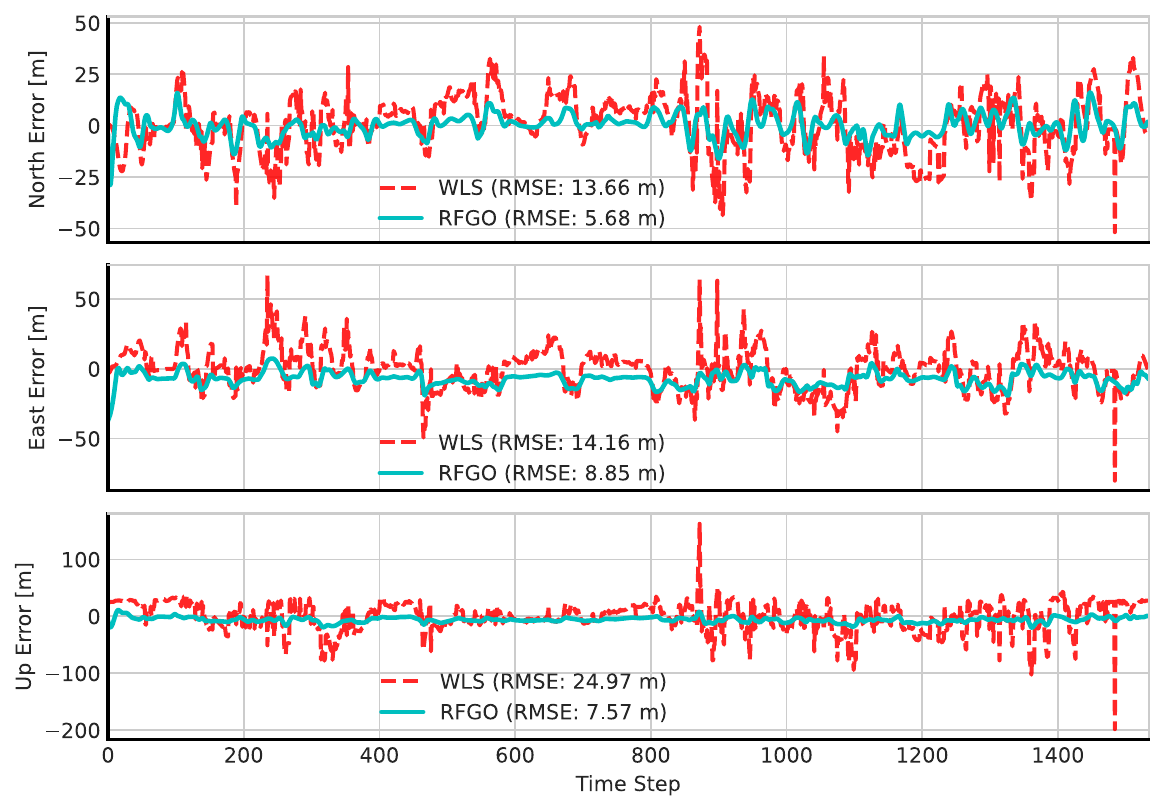}
    \caption{Positioning errors in the East, North, and Up directions over time steps for the WLS and RFGO methods. RMSE values are indicated for each method and direction.}
    \label{fig_error_ENU}
\end{figure}

\begin{figure}[tb!] 
    \centering
    \includegraphics[width=0.81\linewidth]{ 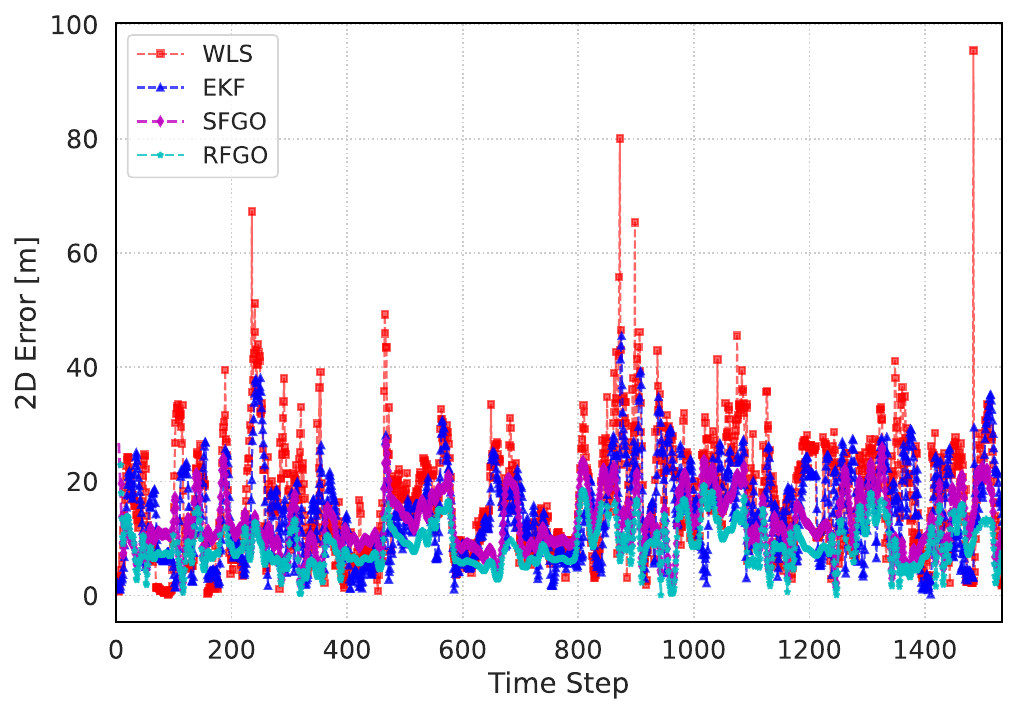}
    \caption{2D positioning errors over time steps for the WLS, EKF, and SFGO methods compared to the RFGO method.}
    \label{fig_2D_error}
\end{figure}

Fig.~\ref{fig_error_ENU} depicts the positioning errors in the East, North, and Up directions over time steps for the proposed RFGO method compared to the WLS baseline.
The RFGO consistently shows reduced error magnitudes across all components, effectively suppressing the large fluctuations observed in WLS. The RMSE values confirm this improvement, with RFGO achieving 5.68 m, 8.85 m, and 7.57 m for the North, East, and Up directions, respectively, compared to 13.66 m, 14.16 m, and 24.97 m for WLS.
Building on this directional error analysis, Fig.~\ref{fig_2D_error} provides a complementary view of the overall 2D positioning errors, comparing RFGO against WLS, EKF, and SFGO over the same time steps.
The proposed RFGO (cyan) achieves the lowest error profile, maintaining values below approximately 20 m for most time steps. The figure indicates that RFGO reduces peak 2D errors by over 50$\%$ relative to SFGO and EKF in high-outlier regions (around time steps 600–800 and 1000–1200), where WLS diverge sharply.
These results imply the enhanced robustness and accuracy of the proposed RFGO framework, highlighting its effectiveness for real-time applications in dynamic and GNSS-challenged settings such as autonomous vehicles.

\begin{figure}[tb!]
    \centering
    \includegraphics[width=0.81\linewidth]{ 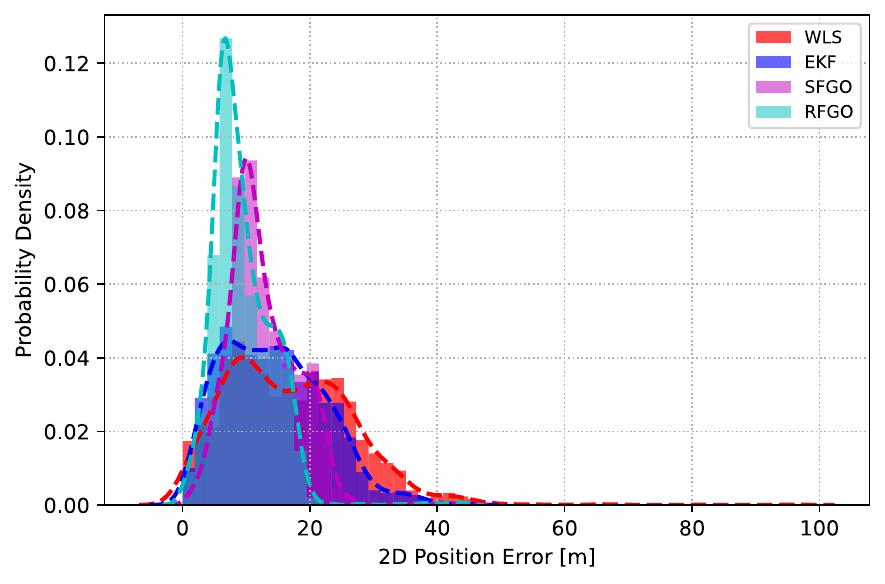}
    \caption{Probability density functions (PDFs) of 2D  positioning errors for different methods. The histograms show the distribution of errors, while the kernel density estimation (KDE) curves provide a smoothed estimate of the underlying probability density.}
    \label{fig_pdf}
\end{figure}

\begin{figure}[tb!]
    \centering
    \includegraphics[width=0.81\linewidth]{ 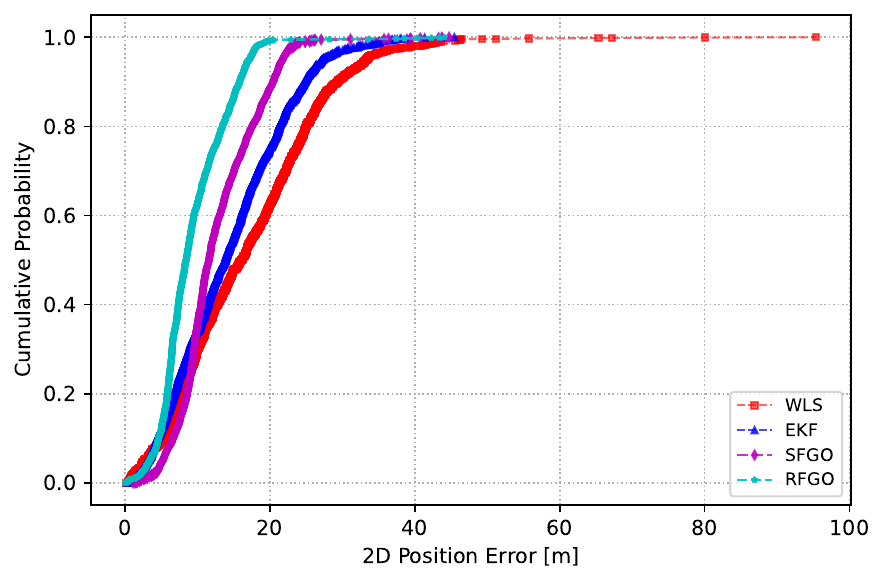}
    \caption{The cumulative distribution functions (CDFs) of the 2D positioning errors for the WLS, EKF, SFGO, and the proposed RFGO methods.}
    \label{fig_cdf}
\end{figure}

To further evaluate the statistical characteristics of the 2D errors, the probability density functions (PDFs) are presented in Fig.~\ref{fig_pdf} for the RFGO method and the baseline approaches: WLS, EKF, and SFGO. The RFGO curve exhibits the sharpest peak near zero error and the quickest decay, indicating a higher concentration of errors within the 0–10 m range compared to the baselines.
Complementing the PDF analysis, Fig.~\ref{fig_cdf} provides the cumulative distribution functions (CDFs) of these errors, offering insight into their progressive accumulation and bounding behavior across methods. The RFGO CDF rises most abruptly, nearing saturation at around 20 m and indicating that all errors remain below this level. In contrast, the SFGO and EKF CDFs reach 90$\%$ cumulative probability at approximately 30 m and 40 m, respectively, while the WLS CDF achieves only 50$\%$ at 40 m, requiring over 100 m for near-complete coverage. From a quantitative perspective, RFGO yields a 95th percentile error of about 15 m, representing a 50–70$\%$ decrease relative to SFGO (30 m), EKF (40 m), and WLS (80 m).

\subsection{Performance Evaluation of RFGO and m-Estimators}
In this subsection, we evaluate the performance of the proposed RFGO with adaptive Barron loss in handling real-world GNSS outliers and enhancing positioning resilience. The evaluation is conducted using the deep part of publicly available UrbanNav dataset. To benchmark the effectiveness, we compare RFGO against existing robust navigation frameworks.
Specifically, we consider traditional FGO approaches that employ fixed m-estimator robust loss functions, including Huber, Cauchy, and Tukey loss functions. These methods are implemented as follows:
\begin{itemize}
    \item FGO-Huber: A TC FGO framework employing the Huber loss function.
    , which combines the properties of $L_2$ and $L_1$ norms.
    \item FGO-Cauchy: A variant of TC FGO that utilizes the Cauchy loss function, which suppresses large residuals more aggressively than Huber.
    \item FGO-Tukey: This approach integrates the Tukey biweight loss into the TC FGO framework, which completely rejects residuals beyond a certain threshold.
\end{itemize}

\begin{figure}[tb!]
    \centering
    \includegraphics[width=0.81\linewidth]{ 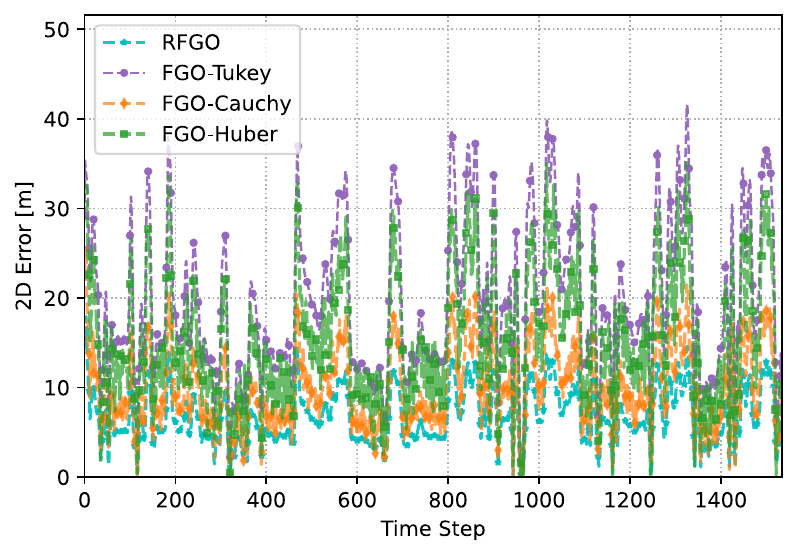}
    \caption{Comparison of 2D positioning error over time steps for different loss functions, RFGO with Barron loss, FGO-Tukey, FGO-Cauchy, and FGO-Huber loss functions in tightly coupled GNSS and IMU integration.}
    \label{fig_error_rfgo_huber_tukey_cachy}
\end{figure}

Figure \ref{fig_error_rfgo_huber_tukey_cachy} depicts the evolution of 2D positioning errors over time steps for the proposed RFGO framework employing the adaptive Barron loss, in comparison to FGO methods utilizing fixed m-estimator robust loss functions: Huber, Tukey, and Cauchy. The figure reveals that the RFGO method maintains lower error magnitudes, while FGO-Huber and FGO-Cauchy show intermediate performance and FGO-Tukey  exhibits the highest fluctuation. This indicates that the adaptive nature of the Barron loss enables more effective outlier mitigation, resulting in smoother error profiles and reduced sensitivity to environmental disturbances compared to the fixed m-estimator approaches.
In addition, Table \ref{tab_error_rfgo_Mestimator} reports the positioning error metrics for RFGO and baseline FGO frameworks with different fixed m-estimator loss functions. The proposed RFGO achieves the lowest values across all metrics, with notable reductions in RMSE of approximately 46$\%$, 53$\%$, and 25$\%$ compared to FGO-Huber, FGO-Tukey, and FGO-Cauchy, respectively.
In addition, the lower SD for RFGO indicates reduced variability, underscoring the adaptive loss function's ability to handle outliers more efficiently than fixed m-estimators. Unlike traditional m-estimators, the Barron loss dynamically adjusts its shape based on residual magnitudes, thereby enhancing both estimation accuracy and robustness under GNSS-challenged environments.
\begin{table}[tb!]
    \centering
    \caption{Comparison of Error Metrics for RFGO with Different m-estimator Robust Loss Functions on the UrbanNav Dataset}
    \label{tab_error_rfgo_Mestimator}
    \renewcommand{\arraystretch}{1.2} 
    \setlength{\tabcolsep}{5pt} 
    \begin{tabular}{l|c|c|c|c}
        \toprule
        \rowcolor[HTML]{FFF2CC}  
        \textbf{Method} & \textbf{ME}$^{\mathrm{a}}$ & \textbf{RMSE}$^{\mathrm{b}}$ & \textbf{MaxE}$^{\mathrm{c}}$ & \textbf{SD}$^{\mathrm{d}}$ \\
        \midrule
        RFGO  & 6.65  & 8.13  & 18.73  & 3.04 \\
        FGO-Huber    & 13.32  & 14.95   & 32.31  & 7.05 \\
        FGO-Tukey    & 15.60  & 17.40   & 39.14  & 8.60 \\
        FGO-Cauchy   & 9.67   & 10.66   & 24.31  & 4.49 \\
        \bottomrule
        \multicolumn{2}{l}{$^{\mathrm{a}}$ Mean 
        Error (MSE).} &
        \multicolumn{3}{l}{$^{\mathrm{b}}$ Root Mean Squared Error (RMSE).} \\
        \multicolumn{2}{l}{$^{\mathrm{c}}$ Max Error (MaxE).} &
        \multicolumn{3}{l}{$^{\mathrm{d}}$ Standard Deviation (SD).} \\
    \end{tabular}
\end{table}

\subsection{Computational Performance Analysis}
In this subsection, we evaluate the computational performance of different estimation strategies for UrbanNav-Deep dataset containing 1535 epochs. Understanding the trade-off between computational efficiency and estimation accuracy is crucial for real-time navigation systems.
The EKF demonstrates the fastest performance, requiring on average 0.008 sec per epoch to process the entire dataset. Despite its speed, EKF's performance comes at the cost of reduced estimation accuracy, as commonly noted in the literature for systems relying on linearized updates and sequential processing.
In contrast, the batch-based RFGO approach significantly increases the computational load, taking on average 0.302 sec per epoch to process the same dataset. This increase is primarily due to the necessity to jointly optimize all variables and measurements, which, while improving accuracy and robustness, demands more intensive computation.

To mitigate this computational overhead, we employed the iSAM2 algorithm. One of the key advantages of iSAM2 is its ability to maintain low computational cost by incrementally updating the solution. Specifically, iSAM2 avoids redundant computations by not recalculating Jacobians for variables that are unaffected by new measurements, leading to improved efficiency while retaining the accuracy of factor graph-based estimation. Accordingly, iSAM2 has a substantially lower computational
time with on average 0.039 sec per epoch to process the dataset.
Moreover, the average runtime per epoch in the RFGO framework nearly doubles under tightly coupled integration compared to loosely coupled integration, due to the increased number of factors in tightly coupled integration. 

\section{Conclusion} \label{conclusion}
This paper introduced a robust tightly coupled GNSS/IMU fusion framework based on an adaptive factor graph optimization framework, RFGO, using the general and adaptive Barron loss function. The proposed RFGO method effectively addresses outliers and non-Gaussian noise, commonly encountered in urban environments, by adaptively tuning the loss function to measurement characteristics. 
Experimental evaluations on the UrbanNav dataset demonstrated a notable reduction in positioning errors, with RFGO achieving a 41$\%$ improvement in RMSE over SFGO, and even greater gains over EKF and WLS. The framework also yielded lower maximum error and variance, particularly under GNSS-challenged conditions.

Future work will focus on reducing computational overhead to support real-time deployment, potentially through parallelization or selective factor updates. Additionally, the current tuning of loss parameters may be replaced with online adaptive mechanisms using reinforcement learning for improved responsiveness. 
Lastly, incorporating complementary sensors such as LiDAR or cameras could further improve performance in GNSS-denied scenarios. 

\section*{Acknowledgment}
This work is funded by the Research Council of Finland–NSF joint project "Distributed AI for enhanced security in satellite-aided wireless navigation - RESILIENT" (359847) and also by the EU–Interreg Aurora project "Enhancing Wireless Communication and Sensing with Secure, Resilient, and Trustworthy Solutions - TRUST". The authors acknowledge the use of ChatGPT for language editing and grammar improvement of the manuscript.

\ifCLASSOPTIONcaptionsoff
  \newpage
\fi

\bibliographystyle{IEEEtran}
\bibliography{references}


\begin{IEEEbiography}
[{\includegraphics[width=1in,height=1.25in,clip,keepaspectratio]{ 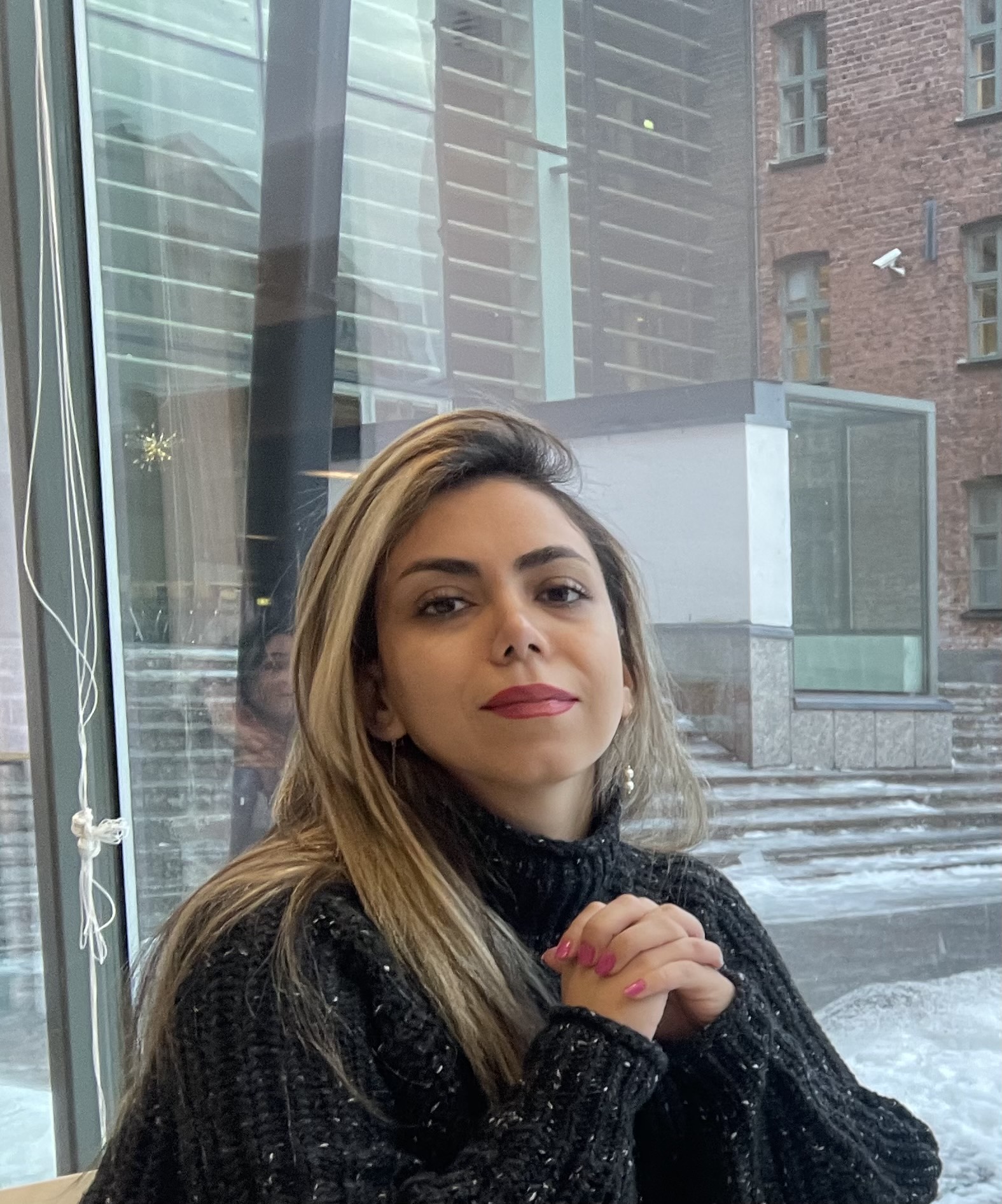}}]
{Elham Ahmadi}
received her M.Sc. degree in electrical and electronics engineering from the Shiraz University of Technology (SUTCH), Iran, in 2017, and Ph.D. degree in automation and system engineering from the Federal University of Santa Catarina (UFSC), Brazil in 2023. She is currently a Postdoctoral researcher at the University of Vaasa, Finland. Her research interests include mathematical optimization, estimation, sensor fusion, control theory, reliable navigation, and machine learning with applications to autonomous systems.
\end{IEEEbiography}

\begin{IEEEbiography}
[{\includegraphics[width=1in,height=1.25in,clip,keepaspectratio]{ 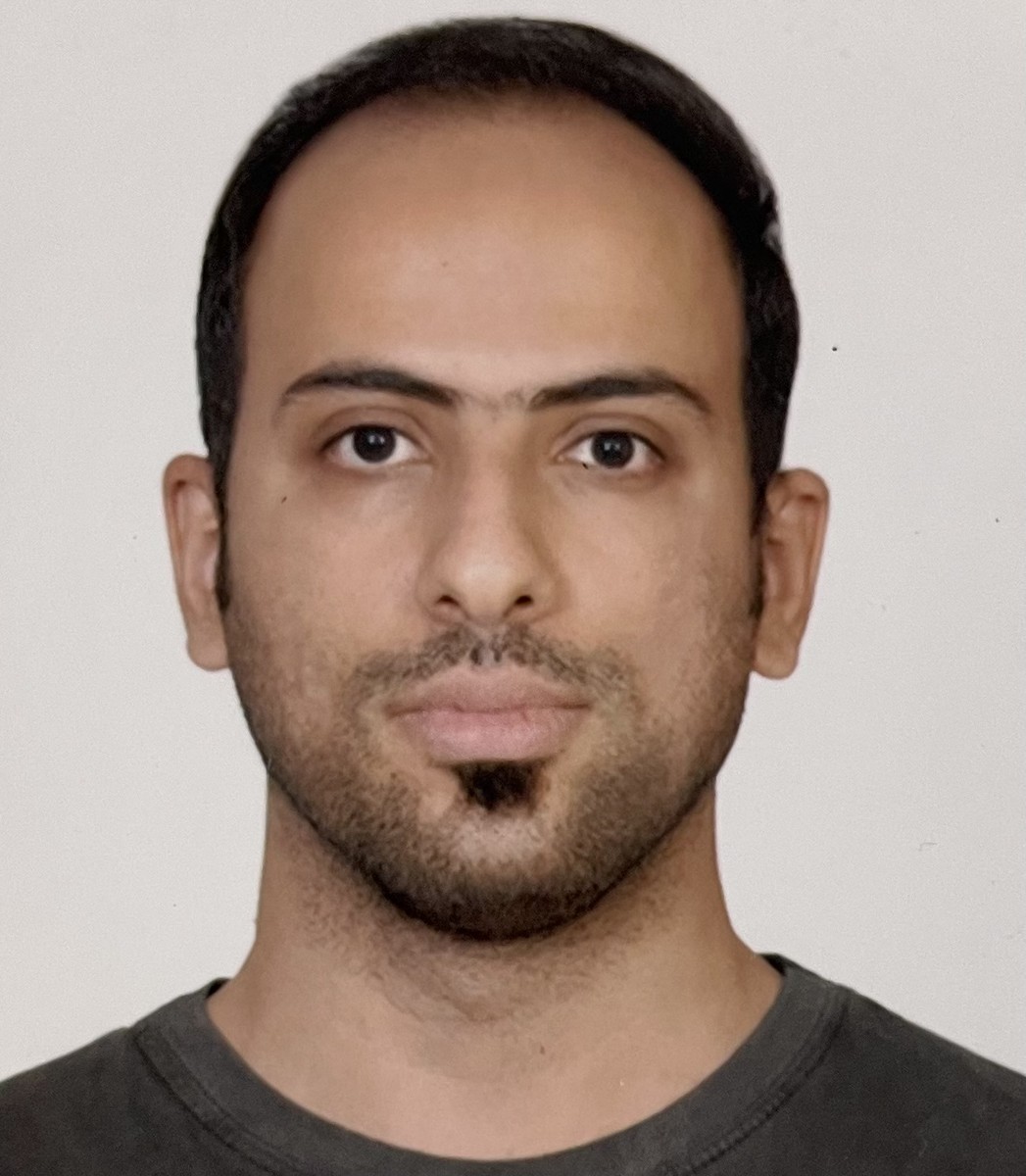}}]
{Alireza Olama}
received his M.Sc. degree in electrical and electronics engineering from the Shiraz University of Technology (SUTCH), Iran, in 2017, and Ph.D. degree in automation and system engineering from the Federal University of Santa Catarina (UFSC), Brazil in 2023. He is currently a Postdoctoral researcher at the Åbo Akademi University, Finland. His research interests include distributed optimization, high-performance computing, machine learning systems, and state estimation.
\end{IEEEbiography}

\begin{IEEEbiography}
[{\includegraphics[width=1in,height=1.25in,clip,keepaspectratio]{ 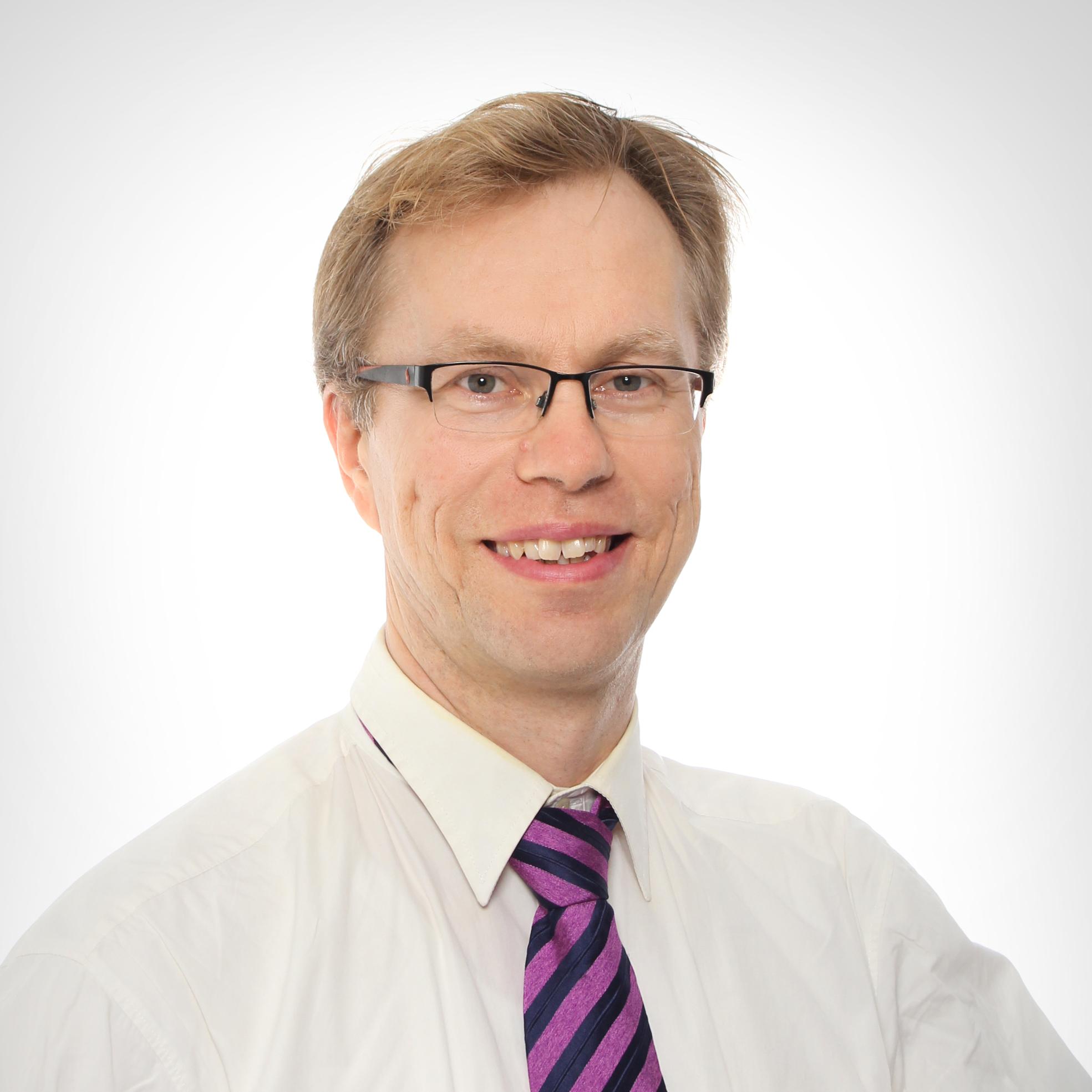}}]
{Petri Välisuo}
received the M.Sc. (Tech.) degree in computer science from the Tampere University of Technology, Tampere, Finland, in 1996, and the D.Sc. (Tech.) degree in automation technology from the University of Vaasa, Vaasa, Finland, in 2011.
He is currently working as a Full Professor, in sustainable automation, at the School of Technology and Innovations, University of Vaasa, Finland.  He has authored and co-authored 40 peer-reviewed publications. His research interests cover machine learning, IoT, positioning methods, and other technologies relevant to industrial automation. He has been working for 10 years in the telecommunication industry before his career at the University of Vaasa.
\end{IEEEbiography}

\begin{IEEEbiography}
[{\includegraphics[width=1in,height=1.25in,clip,keepaspectratio]{ 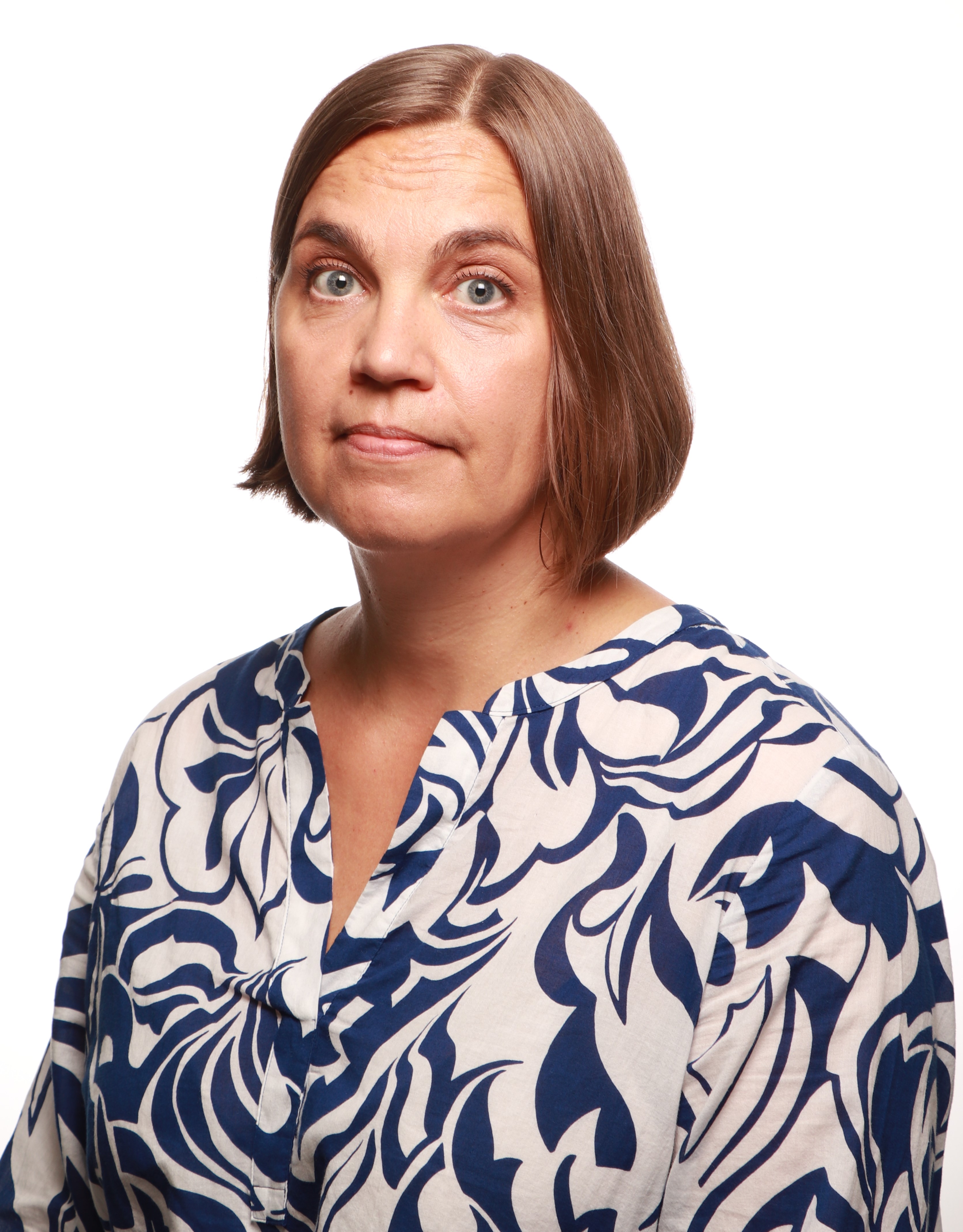}}]
{Heidi Kuusniemi} (Member, IEEE) 
received the M.Sc. (Tech.) degree (with distinction) and the D.Sc. (Tech.) degree from the Tampere University of Technology, Finland, in 2002 and 2005, respectively, both in information technology.
She is a professor in wireless systems at Tampere University and an Adjunct Professor at the University of Vaasa and Aalto University. Her professorship is joint with the Finnish Geospatial Research Institute, National Land Survey of Finland. She is the President of the Nordic Institute of Navigation. She has more than 20 years of experience in research and development of positioning technologies. Her technical expertise and interests include GNSS reliability and resilience, estimation and data fusion, mobile precision positioning, indoor localization and PNT in new space.
\end{IEEEbiography}

\end{document}